\documentclass{article}

\usepackage{microtype}
\usepackage{graphicx}
\usepackage{booktabs} % for professional tables
\usepackage{subfig}

\usepackage{hyperref}

\usepackage[accepted]{icml2025}

\usepackage{amsmath}
\usepackage{amssymb}
\usepackage{mathtools}
\usepackage{amsthm}

\usepackage[capitalize,noabbrev]{cleveref}

\theoremstyle{plain}

\theoremstyle{definition}

\theoremstyle{remark}

\usepackage[textsize=tiny]{todonotes}

\usepackage{multirow}
\usepackage{tikz}
\usetikzlibrary{fit,positioning}
\usepackage{pgfplots}
\usepackage[dvipsnames]{xcolor}

\icmltitlerunning{$\mathcal{X}$-KD: General Experiential Knowledge Distillation for Large Language Models}

\begin{document}

\twocolumn[
\icmltitle{$\mathcal{X}$-KD: General Experiential Knowledge Distillation for Large Language Models}



\icmlsetsymbol{equal}{*}

\begin{icmlauthorlist}
\icmlauthor{Yuang Cai}{yyy}
\icmlauthor{Yuyu Yuan}{yyy}
\end{icmlauthorlist}

\icmlaffiliation{yyy}{Beijing University of Posts and Telecommunications, Beijing, China}

\icmlcorrespondingauthor{Yuang Cai}{cyang@bupt.edu.cn}
\icmlcorrespondingauthor{Yuyu Yuan}{yuanyuyu@bupt.edu.cn}

\icmlkeywords{Machine Learning, ICML}

\vskip 0.3in
]




\begin{abstract}

Knowledge Distillation (KD) for Large Language Models (LLMs) has become increasingly important as models grow in size and complexity.
While existing distillation approaches focus on imitating teacher behavior, they often overlook the original learning environment that shaped the teacher's knowledge.
Inspired by the experiential learning theory and inverse reinforcement learning, we propose Experiential Knowledge Distillation ($\mathcal{X}$-KD), a novel and general framework that enables student models to learn in the teacher's original learning environment.
$\mathcal{X}$-KD adopts the Approximated Variational Reward Imitation Learning (AVRIL) framework to jointly model the teacher's original reward function and perform policy distillation, encouraging consistency between the student policy and the original reward function.
Our derivation demonstrates that $\mathcal{X}$-KD follows the supervised learning framework and applies to both sequence-level and divergence-based distillation methods, underlining the simplicity and flexibility of our approach.
Empirical results show that $\mathcal{X}$-KD outperforms the generalized KD and MiniLLM baselines on abstractive summarization, machine translation, and arithmetic reasoning tasks.
Additionally, $\mathcal{X}$-KD achieves better performance-diversity trade-off and data efficiency than baseline KD approaches.

\end{abstract}

\section{Introduction}

The Experiential Learning Theory \cite{kolb2014experiential} proposes that experiencing the same environment as the original learner can provide the imitative learner with concrete experiences essential for effective learning.
This theory is not only applicable to human learning but also to machine learning.
In machine learning, research has shown that Inverse Reinforcement Learning (IRL) \cite{ng2000algorithms} generally has a better performance compared with Behavioral Cloning (BC) \cite{bain1995framework} in expert imitation learning.
A possible explanation is that in BC, the imitative learner (i.e., the training policy) simply imitates the behaviors of the original learner (i.e., the expert) while in IRL, the imitative learner models the environment (i.e., the reward function) in which the original learner has learned and then places itself to learn in that environment.

The emergence of Large Language Models (LLMs) \cite{brown2020language,raffel2020exploring,openai2022chat,achiam2023gpt} results in the resurgence of knowledge distillation due to the limited energy and computational resources.
Distillation is also a machine learning problem that involves imitative learners (i.e., students) learning from original learners (i.e., teachers). 
The distillation pipelines in the LLM era involve white-box distillation, where the teacher model is accessible, and black-box distillation, where the teacher model is inaccessible.
Applying conventional white-box distillation approaches to LLMs suffers from distribution mismatch, lack of representation ability, and costly teacher-based data generation \cite{kim2016sequence,sanh2019distilbert}.
\citet{wen2023f} propose $f$-Divergence to address distribution mismatch brought by asymmetric distribution divergence.
\citet{liang2023less} propose task-aware distillation, which improves the representation ability of distillation on specific tasks by filtering the useless information in hidden representation.
Recently, GKD \cite{agarwal2024policy} leverages on-policy data and improved divergence to alleviate the issues. 
Black-box distillation is more relevant to prompt engineering and data augmentation, which considers not only the compression of the model but also the more nuanced process of knowledge elicitation and transfer \cite{xu2024survey}.

In knowledge distillation, the teacher model acts as the original learner, while the student model acts as the imitative learner.
Most existing distillation solutions for LLMs, such as use of on-policy data \cite{agarwal2024policy}, improvements on divergence functions \cite{wen2023f}, and filter of hidden representation \cite{liang2023less}, focus mainly on better imitation of the teacher’s behavior without ascertaining the teacher’s original learning environment or putting the student to learn in the original environment.
Likewise, black-box distillation approaches generally put the student model to learn in environments built upon fine-prompted teacher demonstrations or feedbacks \cite{xu2024survey}, which do not directly maximize the reward of the original environment.
None of these approaches put the student model into the teacher’s original learning environment to achieve experiential learning.

Recently, MiniLLM \cite{gu2024minillm}, the concurrent work with the GKD \cite{agarwal2024policy}, proposes to use policy gradient to optimize the reverse KL distillation objective and proves its equivalence with the inverse reinforcement learning (IRL) framework, which considers the reward function of the teacher's original learning environment and breaks through the scope of behavioral cloning.
However, MiniLLM adopts the policy gradient method to maximize a reward function, which deviates from the supervised learning paradigm.
On the other hand, MiniLLM relies on stabilizing techniques to address the training stability issue introduced by the policy gradient method.
These features increase the complexity of integrating MiniLLM into existing LLM training frameworks and therefore may limit the application of MiniLLM's training pipeline in engineering practice.
Furthermore, MiniLLM cannot adapt to other divergence functions or sequence-level distillation for black-box models.

In this work, we propose Experiential Knowledge Distillation ($\mathcal{X}$-KD), a more general and flexible knowledge distillation approach that considers the reward function of the teacher's original learning environment.
We follow Approximated Variational Reward Imitation Learning (AVRIL) \cite{chan2021scalable}, a Bayesian inverse reinforcement learning framework, to integrate experiential learning into sequence-level KD \cite{kim2016sequence}, supervised KD (forward-KL KD) \cite{hinton2015distilling,sanh2019distilbert}, and GKD \cite{agarwal2024policy}.
Specifically, since AVRIL is able to jointly learn the reward function and the policy, we formulate the problem of jointly performing teacher reward learning and sequence-level KD as an AVRIL problem, and then generalize this formulation to forward-KL KD and GKD.
Regarding flexibility, $\mathcal{X}$-KD applies to both sequence-level KD and divergence-based KD and does not limit the type of divergence functions.
Regarding simplicity, the $\mathcal{X}$-KD objective consists of the corresponding original non-experiential KD objective and an experiential regularization term, which follows the supervised training paradigm.
% TODO: experiment result and conclusion
% Experimental results show that $\mathcal{X}$-KD outperforms existing KD methods by delivering better performance, balancing performance and diversity, and improving data efficiency.

Our work makes the following main contributions:
\begin{itemize}
\item We propose $\mathcal{X}$-KD, a novel knowledge distillation framework based on Bayesian inverse reinforcement learning that enables student models to learn in the teacher's original learning environment, rather than just imitating teacher behavior.
\item We demonstrate the flexibility and simplicity of $\mathcal{X}$-KD by integrating it with different knowledge distillation methods while maintaining the supervised learning framework.
\item We empirically demonstrate that $\mathcal{X}$-KD outperforms baseline approaches across multiple NLG tasks while achieving better performance-diversity trade-offs and improved data efficiency.
\end{itemize}

\section{Preliminaries}

\subsection{Knowledge Distillation for LLMs}

Knowledge distillation is generally a problem of training a small student model to transfer the knowledge of a large teacher model.
Conventionally, there are two main types of distillation approaches for LLMs: sequence-level distillation \cite{kim2016sequence} and supervised distillation \cite{hinton2015distilling,sanh2019distilbert}.
Sequence-level distillation maximizes the likelihood of sequences generated by the teacher.
Given a prompt dataset $\mathcal{D}$, the teacher model $\pi$, and the student model $\pi_\theta$, the sequence-level distillation objective can be denoted as Equation \ref{eq:pre-seq-kd}, which can be regarded as SFT on a teacher-generated dataset.
\begin{equation}
\label{eq:pre-seq-kd}
\max_\theta\sum_{\mathbf{x}\in\mathcal{D}}\left[\mathbb{E}_{\mathbf{y}\sim\pi(\cdot|\mathbf{x})}\left[\log\pi_\theta(\mathbf{y}|\mathbf{x})\right]\right]
\end{equation}

Supervised distillation, on the other hand, minimizes the divergence of token-level distributions between the teacher and the student.
Given a supervised fine-tuning (SFT) dataset $\mathcal{D}_\text{SFT}$, the training objective of supervised distillation can be denoted as Equation \ref{eq:pre-skd}.
Here, $D_\text{KL}(\pi||\pi_\theta)(\mathbf{y}|\mathbf{x})$ is the point-wise token-level Kullback-Leibler (KL) divergence with respect to the individual sample $(\mathbf{x},\mathbf{y})$, as shown in Equation \ref{eq:kl-illu}, where $\mathbf{y}_{1:t}=(y_1,y_2,\cdots,y_{t})$ is the token sequence.
\begin{equation}
\label{eq:pre-skd}
\min_\theta\mathbb{E}_{(\mathbf{x},\mathbf{y})\sim\mathcal{D}_\text{SFT}}\left[D_\text{KL}(\pi||\pi_\theta)(\mathbf{y}|\mathbf{x})\right]
\end{equation}
\begin{equation}
\label{eq:kl-illu}
\begin{aligned}
D_\text{KL}&(\pi||\pi_\theta)(\mathbf{y}|\mathbf{x})= \\
&\frac{1}{|\mathbf{y}|}\sum_t D_\text{KL}[\pi(y_{t+1}|\mathbf{x},\mathbf{y}_{1:t})||\pi_\theta(y_{t+1}|\mathbf{x},\mathbf{y}_{1:t})]
\end{aligned}
\end{equation}

Recently, \citet{agarwal2024policy} propose Generalized Knowledge Distillation (GKD) for LLMs, which leverages samples from student models for training and adopts a more generalized divergence function to avoid hallucination and low-quality generations.
The GKD training objective can be denoted as Equation \ref{eq:pre-gkd}.
Here, $D_\beta(\pi||B_\theta)(\mathbf{y}|\mathbf{x})$ denotes the point-wise token-level Jensen-Shannon (JS) divergence \cite{wen2023f} with respect to the individual sample $(\mathbf{x},\mathbf{y})$, as illustrated in Equation \ref{eq:jsd-illu}.
\begin{equation}
\label{eq:pre-gkd}
\min_\theta
\left[
\begin{aligned}
(1-\alpha)\cdot\mathbb{E}_{(\mathbf{x},\mathbf{y})\sim\mathcal{D}_\text{SFT}}\left[D_\beta(\pi||\pi_\theta)(\mathbf{y}|\mathbf{x})\right] \\
+\alpha\cdot\mathbb{E}_{\mathbf{x}\sim\mathcal{D},\mathbf{y}\sim\pi_\theta(\mathbf{y}|\mathbf{x})}\left[D_\beta(\pi||\pi_\theta)(\mathbf{y}|\mathbf{x})\right]
\end{aligned}
\right]
\end{equation}
\begin{equation}
\label{eq:jsd-illu}
\begin{aligned}
D_\beta(\pi||B_\theta)&(\mathbf{y}|\mathbf{x})= \\
&\frac{1}{|\mathbf{y}|}\sum_t D_\beta[\pi(y_{t+1}|\mathbf{x},\mathbf{y}_{1:t})||B_\theta(y_{t+1}|\mathbf{x},\mathbf{y}_{1:t})]
\end{aligned}
\end{equation}

\subsection{MDP Formulation of NLG}

The Natural Language Generation (NLG) task can be formulated as a Markov Decision Process (MDP) \cite{ranzato2016sequence}.
At time step $t$, the state is the prompt along with the previously generated tokens, denoted as $s_t=(\mathbf{x},\mathbf{y}_{1:t})$, and the action is the token to be generated $a_t=y_{t+1}$.
% Note that in auto-regressive decoding, the output tokens are time-shifted.
The action space is the vocabulary $\mathcal{V}$ containing all possible tokens.
% For convenience of later expressions, we regard $\mathcal{V}$ as an ordered set, where $v^{(i)}\in\mathcal{V}$ denotes the $i$-the token in the vocabulary.
In text generation, the state transition is deterministic, so we do not consider the transition probability function.
The reward of taking action $y_{t+1}$ under state $(\mathbf{x},\mathbf{y}_{1:t})$ is denoted as $R(\mathbf{x},\mathbf{y}_{1:t+1})$, which attaches the action to the state for simplicity.
The policy can be denoted as $\pi_\theta(y_{t+1}|\mathbf{x},\mathbf{y}_{1:t})$, which is also the distribution of the language model parameterized by $\theta$.
For simplicity, we sometimes denote the policy as $\pi_\theta(\mathbf{y}|\mathbf{x})=\prod_{t=1}^{|\mathbf{y}|-1}\pi_\theta(y_{t+1}|\mathbf{x},\mathbf{y}_{1:t})$, where $|\mathbf{y}|$ is the length of sequence $|\mathbf{y}|$.
Note that the accumulated product starts from $\pi_\theta(y_2|\mathbf{x},\mathbf{y}_{1:1})$ instead of $\pi_\theta(y_1|\mathbf{x})$ since we assume that all sequences start with a special token denoting the start of the sequence.

\subsection{Bayesian Inverse Reinforcement Learning}

Inverse Reinforcement Learning (IRL) is the problem of extracting a reward function of an MDP given observed optimal behaviors \cite{ng2000algorithms}.
Bayesian Inverse Reinforcement Learning (BIRL) addresses the IRL problem with the Bayesian learning method.
Given the behavioral data $\mathcal{D}$, BIRL formulates the reward distribution as a posterior distribution $p(R|\mathcal{D})$.
The training objective is to approximate the posterior with a parameterized distribution $q_\phi(R)$, i.e., to minimize the KL divergence of $q_\phi(R)$ from $p(R|\mathcal{D})$, as shown in Equation \ref{eq:avril_min_kl}.
Since the posterior is intractable, an equivalent objective is adopted to maximize the Evidence Lower Bound (ELBO) of the KL divergence, as shown in Equation \ref{eq:avril_elbo}.

\begin{equation}
\label{eq:avril_min_kl}
\min_\phi D_\text{KL}[q_\phi(R)||p(R|\mathcal{D})]
\end{equation}
\begin{equation}
\label{eq:avril_elbo}
\max_\phi\mathbb{E}_{R\sim q_\phi(\cdot)}[\log p(\mathcal{D}|R)]-D_\text{KL}[q_\phi(R)||p(R)]
\end{equation}

To practically optimize the ELBO training objective, \cite{chan2021scalable} propose Approximate Variational Reward Imitation Learning (AVRIL), which adopts approximated variational inference to jointly optimize a reward encoder and a Q-value decoder.
The AVRIL training objective is as follows:
\begin{equation}
\label{eq:avril_obj}
\max_{\phi,\theta}\sum_{(s,a,s',a')\in\mathcal{D}}\left[
\begin{aligned}
\log B_\theta(a|s)-D_\text{KL}[q_\phi(\cdot|s,a)||p(\cdot)] \\
+\lambda \log q_\phi(\delta_\theta(s,a,s',a')|s,a)
\end{aligned}
\right]
\end{equation}
Here, $B_\theta(a|s)$ is the Boltzmann policy built upon the Q-value decoder $Q_\theta$ parameterized by $\theta$:
\begin{equation*}
\begin{aligned}
B_\theta(a|s)=\text{softmax}(\beta Q_\theta(s,a))=\frac{\exp(\tau Q_\theta(s,a))}{\sum_{b\in\mathcal{A}}\exp(\tau Q_\theta(s,b))}
\end{aligned}
\end{equation*}
$q_\phi(\cdot|s,a)$ is the current reward distribution given the state $s$ and the action $a$, $p(\cdot)$ is the prior reward distribution, and $\delta_\theta(s,a,s',a')$ is the TD error:
\begin{equation*}
\delta_\theta(s,a,s',s')=Q_\theta(s,a)-\gamma Q_\theta(s',a')
\end{equation*}
The first term trains the Boltzmann policy to maximize the likelihood of the behaviors.
The second term minimizes the KL divergence of the reward posterior from the prior.
However, the likelihood is not conditioned on the reward, which is inconsistent with the ELBO training objective.
The solution is the third term, which forces the TD error computed by the Q-value decoder to follow the reward distribution so that the behaviors are indirectly conditioned on the reward.

\section{Experiential Knowledge Distillation}

\subsection{Original Reward Modeling}
\label{sec:orm}

The objective of experiential learning for knowledge distillation is to enable the student model $\pi_\theta$ to learn in the original environment in which the teacher model $\pi$ has learned.
In other words, given the reward function $R_\pi$ of the teacher's original environment, the objective of experiential learning is to encourage the student model to maximize the expected reward, as shown in Equation \ref{eq:pf-maxr}.
\begin{equation}
\label{eq:pf-maxr}
\max_{\theta}\mathbb{E}_{\mathbf{x}\sim\mathcal{D},\mathbf{y}\sim\pi_\theta(\cdot|\mathbf{x})}[R_\pi(\mathbf{x},\mathbf{y})]
\end{equation}

However, the original reward function $R_\pi$ is generally unknown.
To estimate $R_\pi$, we assume that the original reward function follows a posterior distribution $R_\pi\sim p(R|\pi)$ and adopt a parameterized distribution $q_\phi(R)$ to approximate the posterior.
The intuitive training objective of learning the original reward function is to minimize the KL divergence of $q_\phi(R)$ from $p(R|\pi)$, as shown in Equation \ref{eq:org_reward_min_kl}.
The equivalent objective is to maximize the ELBO of the KL divergence, as shown in Equation \ref{eq:org_reward_elbo}.
\begin{equation}
\label{eq:org_reward_min_kl}
\min_\phi D_\text{KL}[q_\phi(R)||p(R|\pi)]
\end{equation}
\begin{equation}
\label{eq:org_reward_elbo}
\max_\phi\mathbb{E}_{R\sim q_\phi(\cdot)}[\log p(\pi|R)]-D_\text{KL}[q_\phi(R)||p(R)]
\end{equation}
We then approximate the ELBO objective using the AVRIL framework and obtain a practical Original Reward Modeling (ORM) training objective, as shown in Equation \ref{eq:org_reward_avril}.
The derivation is given in Appendix \ref{appendix:derive-orm}.
The ORM training objective is similar to the original AVRIL training objective (Equation \ref{eq:avril_obj}) except that the state-action quadruple $(s,a,s',a')$ is sampled from the teacher policy $\pi$ instead of the offline dataset $\mathcal{D}$, since we model the reward posterior $p(R|\pi)$ conditioning on the teacher policy.
\begin{equation}
\label{eq:org_reward_avril}
\max_{\phi,\theta}
\mathbb{E}_{(s,a,s',a')\sim\pi}\left[
\begin{aligned}
\log\frac{\exp(\tau Q_\theta(s,a))}{\sum_{b\in\mathcal{A}}\exp(\tau Q_\theta(s,b))} \\
-D_\text{KL}[q_\phi(\cdot|s,a)||p(\cdot)] \\
+\lambda\log q_\phi(Q_\theta(s,a)-\gamma Q_\theta(s',a'))
\end{aligned}
\right]
\end{equation}

Regarding the Natural Language Generation (NLG) process as a Markov Decision Process (MDP), given the prompt $\mathbf{x}$, at timestep $t$, the current state is the prompt along with the previously generated response, denoted by $s=(\mathbf{x},\mathbf{y}_{1:t})$, the current action is the currently generated token, denoted by $a=y_{t+1}$, the next state is the prompt along with the currently generated response, denoted by $s'=(\mathbf{x},\mathbf{y}_{1:t+1})$, and the next action is the next token to be generated, denoted by $a'=y_{t+2}$.

% Therefore, under the NLG setting, the sampling of $a$ from policy $\pi$ given the state $s$, which is originally denoted as $a\sim\pi(\cdot|s)$, is now denoted as $y_{t+1}\sim\pi(\cdot|\mathbf{x},\mathbf{y}_{1:t})$, which is the auto-regressive generation given $\pi$ as the language model.
% Furthermore, the expectation over state-action quadruples $\mathbb{E}_{(s,a,s',a')\sim\pi}[f(s,a,s',a')]$ is now denoted as:
% \begin{equation*}
% \mathbb{E}_{\mathbf{x}\sim\mathcal{D}}\mathbb{E}_{\mathbf{y}\sim\pi(\cdot|\mathbf{x})}\sum_{t=1}^{|\mathbf{y}|}f\big(\overbrace{(\mathbf{x},\mathbf{y}_{1:t})}^{s},\underbrace{y_{t+1}}_{a},\overbrace{(\mathbf{x},\mathbf{y}_{1:t+1})}^{s'},\underbrace{y_{t+2}}_{a'}\big)
% \end{equation*}
% where $\mathcal{D}$ is the prompt dataset denoting the initial state distribution.
Therefore, given the prompt dataset $\mathcal{D}$ and the teacher LLM policy $\pi$, we can materialize the ORM training objective (Equation \ref{eq:org_reward_avril}) into the NLG setting by substituting the state-action quadruples with the specific states and actions during auto-regressive generation.
The derivation is given in Appendix \ref{appendix:derive-orm-nlg}.
The loss function of the ORM training objective under the NLG setting is denoted as follows:
\begin{equation}
\label{eq:org_reward_nlg}
\mathcal{L}_\text{orm}(\phi,\theta)=-\sum_{\mathbf{x}\in\mathcal{D}}\mathbb{E}_{\mathbf{y}\sim\pi(\cdot|\mathbf{x})}[\mathcal{F}_\text{avril}(\mathbf{x},\mathbf{y};\phi,\theta)]
\end{equation}
where $\mathcal{F}_\text{avril}$ denotes the AVRIL loss function with respect to the individual sample, as shown in Equation \ref{eq:sample_wise_orm}.
$B_\theta$ is the Boltzmann policy for text generation, as shown in Equation \ref{eq:boltz-nlg}, where $Q_\theta((\mathbf{x},\mathbf{y}_{1:t}),y_{t+1})$ is the Q-value decoder denoting the state-action value of taking action $y_{t+1}$ under state $\mathbf{y}_{1:t}$ and $q_\phi(\cdot|\mathbf{x},\mathbf{y}_{1:t+1})$ is the distribution of the reward after taking action $a_t=y_{t+1}$ and transferring to state $s_{t+1}=\mathbf{y}_{1:t+1}$.
$\delta_t(\theta)$ denotes the TD error at timestep $t$, i.e., the difference between the Q-value of the current action $y_{t+1}$ and the discounted Q-value of the next action $y_{t+2}$, as shown in Equation \ref{eq:td-nlg}.
\begin{equation}
\label{eq:sample_wise_orm}
\mathcal{F}_\text{avril}(\mathbf{x},\mathbf{y};\phi,\theta)=\sum_{t=1}^{|\mathbf{y}|}\left[
\begin{aligned}
\log B_\theta(y_{t+1}|\mathbf{x},\mathbf{y}_{1:t}) \\
-D_\text{KL}[q_\phi(\cdot|\mathbf{x},\mathbf{y}_{1:t+1})||p(\cdot)] \\
+\lambda \log q_\phi(\delta_t(\theta)|\mathbf{x},\mathbf{y}_{1:t+1})
\end{aligned}
\right]
\end{equation}
\begin{equation}
\label{eq:boltz-nlg}
B_\theta(y_{t+1}|\mathbf{x},\mathbf{y}_{1:t}) = \frac{\exp(\tau Q_\theta((\mathbf{x},\mathbf{y}_{1:t}),y_{t+1}))}{\sum_{y'\in\mathcal{V}}\exp(\tau Q_\theta((\mathbf{x},\mathbf{y}_{1:t}),y'))}
\end{equation}
\begin{equation}
\label{eq:td-nlg}
\delta_t(\theta) = Q_\theta((\mathbf{x},\mathbf{y}_{1:t}),y_{t+1}) - \gamma Q_\theta((\mathbf{x},\mathbf{y}_{1:t+1}),y_{t+2})
\end{equation}
Note that we replace the expectation expression $\mathbb{E}_{\mathbf{x}\sim\mathcal{D}}$ by the summation expression $\sum_{\mathbf{x}\in\mathcal{D}}$, which does not affect the optimality of the training objective.

\subsection{Sequence-Level $\mathcal{X}$-KD}

Regarding the Boltzmann policy $B_\theta$ as the student policy, the ORM objective (Equation \ref{eq:org_reward_nlg}) can be reformulated as the combination of a knowledge distillation objective and an experiential objective:
\begin{equation}
\label{eq:seq-x}
\begin{aligned}
\mathcal{L}_\text{orm}(\phi,\theta)=\mathcal{L}_\text{seq}(\theta)+\mathcal{L}_\text{ex}(\phi,\theta)
\end{aligned}
\end{equation}
Here, $\mathcal{L}_\text{seq}$ is the sequence-level knowledge distillation (KD) objective \cite{kim2016sequence} which distills the knowledge in the teacher policy $\pi$ into the student policy $B_\theta$:
\begin{equation}
\label{eq:seqkd}
\begin{aligned}
\mathcal{L}_\text{seq}(\theta)&=-\sum_{\mathbf{x}\in\mathcal{D}}\mathbb{E}_{\mathbf{y}\sim\pi(\cdot|\mathbf{x})}\left[\log B_\theta(\mathbf{y}|\mathbf{x})\right] \\
&=-\sum_{\mathbf{x}\in\mathcal{D}}\mathbb{E}_{\mathbf{y}\sim\pi(\cdot|\mathbf{x})}\left[\sum_{t=1}^{|\mathbf{y}|}\log B_\theta(y_{t+1}|\mathbf{x},\mathbf{y}_{1:t})\right]
\end{aligned}
\end{equation}
$\mathcal{L}_\text{ex}$ is the experiential objective ensuring the reward distribution to satisfy the prior distribution $p(R)$ and ensuring the student policy $B_\theta$ to satisfy the TD-error constraint under the reward distribution:
\begin{equation}
\mathcal{L}_\text{ex}(\phi,\theta)=\sum_{\mathbf{x}\in\mathcal{D}}\mathbb{E}_{\mathbf{y}\sim\pi(\cdot|\mathbf{x})}\left[\mathcal{F}_\text{ex}(\mathbf{x},\mathbf{y};\phi,\theta)\right]
\end{equation}
where $\mathcal{F}_\text{ex}$ is the sample-wise experiential objective:
\begin{equation}
\label{eq:fex}
\mathcal{F}_\text{ex}(\mathbf{x},\mathbf{y};\phi,\theta)=\sum_{t=1}^{|\mathbf{y}|}\left[
\begin{aligned}
D_\text{KL}[q_\phi(\cdot|\mathbf{x},\mathbf{y}_{1:t+1})||p(\cdot)] \\
-\lambda\log q_\phi(\delta_t(\theta)|\mathbf{x},\mathbf{y}_{1:t+1})
\end{aligned}
\right]
\end{equation}
Here, the hyperparameter $\lambda$ is referred to as the experiential weight since it directly controls the degree of TD-error constraint.

The loss $\mathcal{L}_\text{orm}$ can also be referred to as the sequence-level \textbf{Ex}periential \textbf{K}nowledge \textbf{D}istillation (sequence-level $\mathcal{X}$-KD) loss.
By minimizing $\mathcal{L}_\text{orm}$, the student policy $B_\theta$ performs sequence-level knowledge distillation from the teacher policy $\pi$ through the minimization of $\mathcal{L}_\text{seq}$, while staying consistent with the reward function of the teacher's original learning environment through the minimization of $\mathcal{L}_\text{ex}$, thus achieving experiential learning.

\subsection{Supervised $\mathcal{X}$-KD}
\label{sec:s-xkd}

The sequence-level KD objective $\mathcal{L}_\text{seq}$ can further be written as Equation \ref{eq:seq-reform}, where the first term in the square brackets is the KL divergence of the teacher policy $\pi$ from the student policy $B_\theta$.
The derivation is given in Appendix \ref{appendix:derive-seqkd}.
Note that the second term is independent of the parameter $\theta$, which can be ignored during optimization.
In other words, minimizing the KL divergence is equivalent to minimizing $\mathcal{L}_\text{seq}$.
\begin{equation}
\label{eq:seq-reform}
\mathcal{L}_\text{seq}(\theta)=\sum_{\mathbf{x}\in\mathcal{D}}
\left[
\begin{aligned}
D_\text{KL}[\pi(\cdot|\mathbf{x})||B_\theta(\cdot|\mathbf{x})] \\
-\mathbb{E}_{\mathbf{y}\sim\pi(\cdot|\mathbf{x})}[\log \pi(\mathbf{y}|\mathbf{x})]
\end{aligned}
\right]
\end{equation}
Therefore, we combine the KL divergence with the experiential objective to obtain the objective in Equation \ref{eq:seq-x-div}.
\begin{equation}
\label{eq:seq-x-div}
\begin{aligned}
\mathcal{L}_\text{xkd}(\phi,\theta)&=\mathcal{L}_\text{kd}(\theta)+\mathcal{L}_\text{ex}(\phi,\theta) \\
&=\sum_{\mathbf{x}\in\mathcal{D}}D_\text{KL}[\pi(\cdot|\mathbf{x})||B_\theta(\cdot|\mathbf{x})]
+\mathcal{L}_\text{ex}(\phi,\theta)
\end{aligned}
\end{equation}

The loss $\mathcal{L}_\text{xkd}$ is referred to as the supervised $\mathcal{X}$-KD loss.
By minimizing $\mathcal{L}_\text{xkd}$, the student policy $B_\theta$ performs supervised knowledge distillation from the teacher policy $\pi$ through the minimization of $\mathcal{L}_\text{kd}$, while staying consistent with the reward function of the teacher's original learning environment through the minimization of $\mathcal{L}_\text{ex}$, thus achieving experiential learning.

\subsection{Generalized $\mathcal{X}$-KD}
\label{sec:g-xkd}

The sequence-level KD loss in Eq. \ref{eq:seqkd} denotes a forward sequence-level KD objective, which encourages the student model to maximize the output probabilities of sentences generated by the teacher model.
The corresponding reverse sequence-level KD objective is to encourage the student model to generate sentences that have higher probabilities under the teacher model, as shown in Eq. \ref{eq:seqkd-rev}.
A more general sequence-level KD objective can be denoted as the combination of the forward and the reverse objective, as shown in Eq. \ref{eq:seqkd-gen}, where $\beta$ is the hyperparameter balancing the two objectives.
\begin{align}
\label{eq:seqkd-rev}
\mathcal{L}_\text{r-seq}(\theta)&=-\sum_{\mathbf{x}\in\mathcal{D}}\mathbb{E}_{\mathbf{y}\sim B_\theta(\cdot|\mathbf{x})}\left[\log \pi(\mathbf{y}|\mathbf{x})\right] \\
\label{eq:seqkd-gen}
\mathcal{L}_\text{g-seq}(\theta)&=\beta\cdot\mathcal{L}_\text{seq}(\theta)+(1-\beta)\cdot\mathcal{L}_\text{r-seq}(\theta)
\end{align}

In supervised $\mathcal{X}$-KD, we reformulate $\mathcal{L}_\text{seq}$ as the combination of KL divergence and negative log-likelihood, as shown in Eq. \ref{eq:seq-reform}.
The derivation is given in Appendix \ref{appendix:derive-seqkd}
The motivation is to obtain the conventional KL-based knowledge distillation loss $\mathcal{L}_\text{kd}$ through the KL divergence.
Similarly, we reformulate the general sequence-level KD loss $\mathcal{L}_\text{g-seq}$ as the combination of $\beta$-JS divergence and negative log-likelihood, as shown in Eq. \ref{eq:gen-seq-reform}.
\begin{equation}
\label{eq:gen-seq-reform}
\mathcal{L}_\text{g-seq}(\theta)=\sum_{\mathbf{x}\in\mathcal{D}}
\left[
\begin{aligned}
D_\beta[\pi(\cdot|\mathbf{x})||B_\theta(\cdot|\mathbf{x})] \\
-\beta\mathbb{E}_{\mathbf{y}\sim\pi(\cdot|\mathbf{x})}[\log \pi(\mathbf{y}|\mathbf{x})] \\
-(1-\beta)\mathbb{E}_{\mathbf{y}\sim B_\theta(\cdot|\mathbf{x})}[\log B_\theta(\mathbf{y}|\mathbf{x})]
\end{aligned}
\right]
\end{equation}
We then combine $\mathcal{L}_\text{g-seq}$ with the experiential loss $\mathcal{L}_\text{ex}$ and remove unuseful terms to obtain the training objective in Eq. \ref{eq:g-xkd}.
\begin{equation}
\label{eq:g-xkd}
\begin{aligned}
\mathcal{L}_\text{g-xkd}(\phi,\theta)&=\mathcal{L}_\text{gkd}(\theta)+\mathcal{L}_\text{ex}(\phi,\theta) \\
&=\sum_{\mathbf{x}\in\mathcal{D}}D_\beta[\pi(\cdot|\mathbf{x})||B_\theta(\cdot|\mathbf{x})]
+\mathcal{L}_\text{ex}(\phi,\theta)
\end{aligned}
\end{equation}
Here, $\mathcal{L}_\text{gkd}(\theta)$ denotes the GKD loss, where the $\beta$-JS divergence is estimated using the mixture of on-policy and offline data, as shown in Eq. \ref{eq:gkd}, with $\alpha$ denoting the ratio of on-policy data.
\begin{equation}
\label{eq:gkd}
\begin{aligned}
\mathcal{L}_\text{gkd}(\theta)=(1-\alpha)\cdot\mathbb{E}_{(\mathbf{x},\mathbf{y})\in\mathcal{D}_\text{SFT}}\left[D_\beta(\pi||B_\theta)(\mathbf{y}|\mathbf{x})\right] \\
+\alpha\cdot\mathbb{E}_{\mathbf{x}\sim\mathcal{D},\mathbf{y}\sim B_\theta(\cdot|\mathbf{x})}\left[D_\beta(\pi||B_\theta)(\mathbf{y}|\mathbf{x})\right]
\end{aligned}
\end{equation}

The loss $\mathcal{L}_\text{g-xkd}$ is referred to as the generalized $\mathcal{X}$-KD loss.
By minimizing $\mathcal{L}_\text{g-xkd}$, the student policy $B_\theta$ not only distills knowledge from the teacher policy $\pi$ through the minimization of the GKD loss $\mathcal{L}_\text{gkd}$, but also stays consistent with the reward function of the teacher's original learning environment through the minimization of $\mathcal{L}_\text{ex}$, thus achieving experiential learning.
Note that the loss $\mathcal{L}_\text{g-xkd}$ is based on the assumption that the loss $\mathcal{L}_\text{seq}$ in Equation \ref{eq:seq-x} can be replaced by any other sequence-level KD loss.

\begin{algorithm}[tb]
\caption{Generalized $\mathcal{X}$-KD.}
\label{alg:distill}
\begin{algorithmic}[1]
\STATE {\bfseries Input:} prompt dataset $\mathcal{D}$, SFT dataset $\mathcal{D}_\text{SFT}$, teacher policy $\pi$, student policy $p_\theta$, reward distribution $q_\phi$
\STATE {\bfseries Params:}  on-policy weight $\alpha$, experiential weight $\lambda$
\STATE {\bfseries Output:} new student policy $p_\theta$
\STATE Fine-tune $\pi$ and $p_\theta$ with $\mathcal{D}_\text{SFT}$
\FOR{$k=1,\cdots,T_1$}
\STATE Sample $u\sim\text{Uniform}(0,1)$
\IF{$u\leq\alpha$}
\STATE Sample $\mathbf{x}\sim\mathcal{D}$, and sample $\mathbf{y}\sim p_\theta(\cdot|\mathbf{x})$
\ELSE
\STATE Sample $(\mathbf{x},\mathbf{y})\sim\mathcal{D}_\text{SFT}$
\ENDIF
\STATE $\phi,\theta\leftarrow\nabla_{\phi,\theta}\left[D_\beta(\pi||p_\theta)(\mathbf{y}|\mathbf{x})+\mathcal{F}_\text{ex}(\mathbf{x},\mathbf{y};\phi,\theta)\right]$
\ENDFOR
\STATE {\bfseries return} $p_\theta$
\end{algorithmic}
\end{algorithm}

\begin{algorithm}[tb]
\caption{Black-box (sequence-level) $\mathcal{X}$-KD.}
\label{alg:blackbox-distill}
\begin{algorithmic}[1]
\STATE {\bfseries Input:} prompt dataset $\mathcal{D}$, black-box teacher policy $\pi^*$, student policy $p_\theta$, reward distribution $q_\phi$
\STATE {\bfseries Params:}  experiential weight $\lambda$
\STATE {\bfseries Output:} new student policy $p_\theta$
\STATE Fine-tune $p_\theta$ with $\mathcal{D}_\text{SFT}$
\FOR{$k=1,\cdots,T_1$}
\STATE Sample $u\sim\text{Uniform}(0,1)$
\STATE Sample $\mathbf{x}\sim\mathcal{D}$, and sample $\mathbf{y}\sim \pi^*(\cdot|\mathbf{x})$
\STATE $\phi,\theta\leftarrow\nabla_{\phi,\theta}\left[\log p_\theta(\mathbf{y}|\mathbf{x})+\mathcal{F}_\text{ex}(\mathbf{x},\mathbf{y};\phi,\theta)\right]$
\ENDFOR
\STATE {\bfseries return} $p_\theta$
\end{algorithmic}
\end{algorithm}

\subsection{$\mathcal{X}$-KD Pipelines}

% TODO: Involve the initialization of the Boltzmann policy.
In the $\mathcal{X}$-KD formulation, the student policy $B_\theta$ is denoted by the Q-value decoder $Q_\theta$, and the computation of the $\mathcal{X}$-KD loss depends on $Q_\theta$.
However, in language model distillation, the student policy is generally initialized with a pre-trained language policy $p_\theta$, without an explicit definition of the Q-value model.
To compute the $\mathcal{X}$-KD objective through $p_\theta$, we inversely formulate the Q-value decoder $Q_\theta$ as the log-softmax function of action probabilities, as shown in Equation \ref{eq:inverse-boltz}, which is a non-strict inversion of the Boltzmann policy formulation.
\begin{equation}
\label{eq:inverse-boltz}
Q_\theta((\mathbf{x},\mathbf{y}_{1:t}),y_{t+1})=\log\frac{\exp(\tau'\cdot p_\theta(y_{t+1}|\mathbf{x},\mathbf{y}_{1:t}))}{\sum_{\mathbf{y}'\in\mathcal{V}}\exp(\tau'\cdot p_\theta(v|\mathbf{x},\mathbf{y}_{1:t}))}
\end{equation}

Given the teacher policy $\pi$ and the student policy $p_\theta$, the pipeline of generalized $\mathcal{X}$-KD is shown in Algorithm \ref{alg:distill}.
Here, the computation of the sample-wise experiential objective $\mathcal{F}_\text{ex}(\mathbf{x},\mathbf{y};\phi,\theta)$ depends on the TD-error $\delta_t(\theta)$, as shown in Equation \ref{eq:fex}, where the experiential weight $\lambda$ works.
And the computation of the TD-error further depends on the Q-value function $Q_\theta$, which is now substituted by Equation \ref{eq:inverse-boltz}.
The pipeline degenerates into supervised $\mathcal{X}$-KD when setting the on-policy weight $\alpha$ to 0, and degenerates into conventional non-experiential GKD when setting the experiential weight $\lambda$ to 0.

The above pipeline only applies to white-box distillation where the output distribution of the teacher model is accessible.
However, for some powerful language models, we only have access to their final output responses.
Therefore, in addition to white-box $\mathcal{X}$-KD, we also directly leverage the sequence-level $\mathcal{X}$-KD objective (Equation \ref{eq:seq-x}) for black-box distillation, as shown in Algorithm \ref{alg:blackbox-distill}.
It is worth noting that Algorithm \ref{alg:blackbox-distill} also applies to white-box models as a sequence-level experiential distillation approach.
The pipeline degenerates into conventional sequence-level distillation when setting the experiential weight $\lambda$ to 0.

\section{Experiment}

\begin{table}[H]
\setlength\tabcolsep{2pt}
\centering
\caption{Overall results of $\mathcal{X}$-KD and baseline approaches on multiple NLG tasks with different model initializations.}
\label{tab:overall-results}
\begin{tabular}{@{}lrrrrrr@{}}
\toprule
\multicolumn{1}{c}{} & \multicolumn{3}{c}{\textbf{large$\rightarrow$small}}                                                 & \multicolumn{3}{c}{\textbf{large$\rightarrow$base}}                                                 \\
\multicolumn{1}{c}{} & \multicolumn{1}{c}{XSum} & \multicolumn{1}{c}{WMT} & \multicolumn{1}{c}{GSM} & \multicolumn{1}{c}{XSum} & \multicolumn{1}{c}{WMT} & \multicolumn{1}{c}{GSM} \\ \midrule
\textit{Baselines}   &                          &                         &                         &                          &                         &                         \\
SFT                  & 12.78                    & 26.04                   & 8.33                    & 15.90                    & 27.22                   & 11.21                   \\
SeqKD                & 13.30                    & 26.15                   & 8.94                    & 16.07                    & 27.26                   & 12.05                   \\
SKD                  & 13.38                    & 26.10                   & 9.17                    & 16.10                    & 27.24                   & 11.97                   \\
GKD                  & 13.62                    & 26.61                   & 12.35                   & 16.18                    & 27.43                   & 13.48                   \\
MiniLLM              & 13.69                    & 26.65                   & \textbf{12.50}          & 16.15                    & 27.54                   & 13.41                   \\ \midrule
\textit{Ours}        &                          &                         &                         &                          &                         &                         \\
SeqXKD               & 13.36                    & 26.23                   & 9.32                    & 16.09                    & 27.28                   & 12.42                   \\
S-XKD                & 13.45                    & 26.63                   & 10.53                   & 16.12                    & 27.44                   & 12.50                   \\
G-XKD                & \textbf{13.81}           & \textbf{26.73}          & 12.35                   & \textbf{16.24}           & \textbf{27.61}          & \textbf{13.63}          \\ \bottomrule
\end{tabular}
\end{table}

\subsection{Experiment Setup}

\textbf{Tasks and Datasets.} We perform knowledge distillation on abstractive summarization, machine translation, and arithmetic reasoning tasks.
For abstractive summarization, we use the XSum \cite{narayan2018don} dataset containing news articles paired with single-sentence summaries.
We evaluate the summarization performance on the validation set and report the ROUGE scores.
For machine translation, we perform the English-to-German (en-de) translation task using the top 10k training data from the WMT14 \cite{bojar2014findings} de-en subset.
We evaluate the translation performance on the validation set and report the BLEU scores.
For arithmetic reasoning, we use the GSM8k \cite{cobbe2021training} dataset containing grade school math word problems paired with detailed solutions.
We use the same few-shot prompting as in GKD to improve the arithmetic reasoning performance.
We evaluate the arithmetic reasoning performance on the test set and report the accuracy scores.

\textbf{Model Initialization.} For white-box distillation, we initialize the teacher and student with T5 model \cite{raffel2020exploring} of different sizes.
Specifically, we initialize the teacher models with T5-large (780M) and initialize the student models with T5-small (77M) and T5-base (250M).
For black-box distillation, due to the expensive cost of invoking powerful black-box models, we employ Llama-2-7b as the black-box teacher model and adopt T5-large as the student model.
For white-box teacher models and all student models, we perform supervised fine-tuning on specific tasks before knowledge distillation.

\textbf{Baselines.} We compare the $\mathcal{X}$-KD approaches with the baseline KD approaches that do not incorporate experiential learning.
Specifically, for sequence-level or black-box distillation, we compare sequence-level $\mathcal{X}$-KD with the conventional sequence-level KD approach introduced in \citet{kim2016sequence}.
For divergence-based distillation, we compare supervised $\mathcal{X}$-KD with the conventional supervised KD approach introduced in \citet{hinton2015distilling,sanh2019distilbert}, and compare generalized $\mathcal{X}$-KD with GKD \cite{agarwal2024policy}.
The hyperparameter configurations of various approaches are given in Appendix \ref{appendix:hyperparam}.

\textbf{Performance-Diversity.} To explore the performance-diversity trade-off abilities of $\mathcal{X}$-KD, we adjust the sampling temperature to generate results with different diversities on the evaluation dataset.
Specifically, we use 4 different sampling temperatures, which are 0.1, 0.3, 0.5, and 1.0.
We then compute the SelfBLEU scores of different evaluation results to measure their diversities.
A lower SelfBLEU score indicates more diverse outputs, therefore, we use 1 - SelfBLEU as the diversity score.
Finally, we plot how performance scores vary with changes in the SelfBLEU score.

\textbf{Data Efficiency Study.} We perform data efficiency studies on all three NLG tasks to explore whether $\mathcal{X}$-KD can achieve comparable performance to baseline approaches with less distillation data.
Specifically, we train baseline KD and $\mathcal{X}$-KD on 25\%, 50\%, 75\%, and 100\% of the training set.
Here, we use T5-large as the teacher model and T5-small as the student model.
The settings of training steps decrease proportionally with the training set.

Additional experiment results are included in Appendix \ref{appendix:addexp}.
More detailed experimental settings (e.g., hyperparameters) are included in Appendix \ref{appendix:hyperparam}.

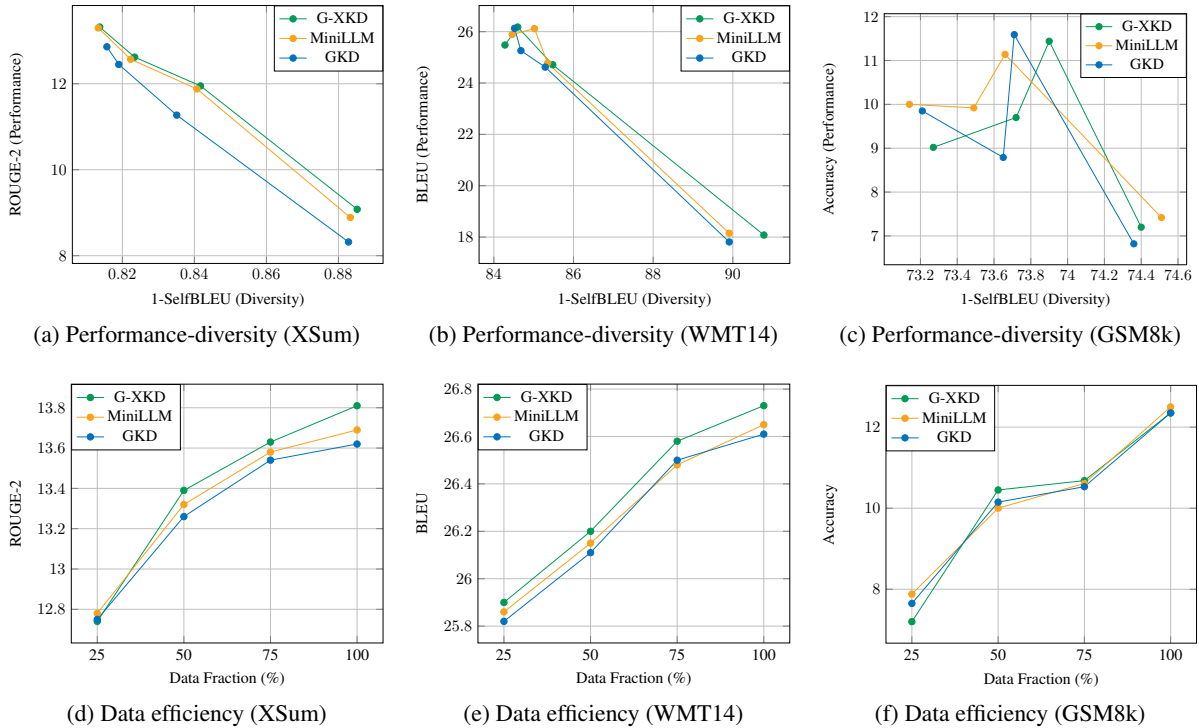
\begin{figure*}
\centering
% xsum diversity
\subfloat[Performance-diversity (XSum)]{
    \centering
    \resizebox{0.3\textwidth}{!}{
    \begin{tikzpicture}
    \begin{axis}[
        grid=major,
        xlabel=1-SelfBLEU (Diversity),
        ylabel=ROUGE-2 (Performance),
        legend style={at={(1,1)},anchor=north east}
    ]
    \addplot[mark=*,ForestGreen] plot coordinates {
        (0.8137,13.32) 
        (0.8234,12.62) 
        (0.8417,11.95) 
        (0.8852,9.08) 
    };
    \addlegendentry{G-XKD}
    \addplot[mark=*,YellowOrange] plot coordinates {
        (0.8133,13.30) 
        (0.8223,12.57)
        (0.8407,11.88)
        (0.8833,8.89)
    };
    \addlegendentry{MiniLLM}
    \addplot[mark=*,RoyalBlue] plot coordinates {
        (0.8157,12.86) 
        (0.8190,12.45)
        (0.8351,11.27) 
        (0.8828,8.32)
    };
    \addlegendentry{GKD}
    \end{axis}
    \end{tikzpicture}
    }
    \label{fig:xsum-p-d}
}
% wmt diversity
\subfloat[Performance-diversity (WMT14)]{
    \centering
    \resizebox{0.3\textwidth}{!}{
    \begin{tikzpicture}
    \begin{axis}[
        grid=major,
        xlabel=1-SelfBLEU (Diversity),
        ylabel=BLEU (Performance),
        legend style={at={(1,1)},anchor=north east}
    ]
    \addplot[mark=*,ForestGreen] plot coordinates {
        (84.28,25.48)
        (84.60,26.18)
        (85.48,24.71)
        (90.78,18.08)
    };
    \addlegendentry{G-XKD}
    \addplot[mark=*,YellowOrange] plot coordinates {
        (84.46,25.89)
        (85.02,26.12)
        (85.35,24.78)
        (89.91,18.15)
    };
    \addlegendentry{MiniLLM}
    \addplot[mark=*,RoyalBlue] plot coordinates {
        (84.52,26.13)
        (84.68,25.26)
        (85.29,24.62)
        (89.91,17.81)
    };
    \addlegendentry{GKD}
    \end{axis}
    \end{tikzpicture}
    }
    \label{fig:wmt-p-d}
}
% gsm diversity
\subfloat[Performance-diversity (GSM8k)]{
    \centering
    \resizebox{0.3\textwidth}{!}{
    \begin{tikzpicture}
    \begin{axis}[
        grid=major,
        xlabel=1-SelfBLEU (Diversity),
        ylabel=Accuracy (Performance),
        legend style={at={(1,1)},anchor=north east}
    ]
    \addplot[mark=*,ForestGreen] plot coordinates {
        (73.27,9.02)
        (73.72,9.70)
        (73.90,11.44)
        (74.40,7.20)
    };
    \addlegendentry{G-XKD}
    \addplot[mark=*,YellowOrange] plot coordinates {
        (73.14,10.00)
        (73.49,9.92)
        (73.66,11.14)
        (74.51,7.42)
    };
    \addlegendentry{MiniLLM}
    \addplot[mark=*,RoyalBlue] plot coordinates {
        (73.21,9.85)
        (73.65,8.79)
        (73.71,11.59)
        (74.36,6.82)
    };
    \addlegendentry{GKD}
    \end{axis}
    \end{tikzpicture}
    }
    \label{fig:gsm-p-d}
}
\hfill
% xsum data-eff
\subfloat[Data efficiency (XSum)]{
\centering
\resizebox{0.3\textwidth}{!}{
    \begin{tikzpicture}
    \begin{axis}[
        grid=major,
        xlabel=Data Fraction (\%),
        ylabel=ROUGE-2,
        symbolic x coords={25,50,75,100},
        xtick=data,
        xticklabel style={name=tick no \ticknum},
        legend style={at={(0,1)},anchor=north west}
    ]
    \addplot[mark=*,ForestGreen] plot coordinates {
        (25, 12.74)
        (50, 13.39)
        (75, 13.63)
        (100,13.81)
    };
    \addlegendentry{G-XKD}
    \addplot[mark=*,YellowOrange] plot coordinates {
        (25, 12.78)
        (50, 13.32)
        (75, 13.58)
        (100,13.69)
    };
    \addlegendentry{MiniLLM}
    \addplot[mark=*,RoyalBlue] plot coordinates {
        (25, 12.75)
        (50, 13.26)
        (75, 13.54)
        (100,13.62)
    };
    \addlegendentry{GKD}
    \end{axis}
    \end{tikzpicture}
    }
    \label{fig:xsum-data-eff}
}
% wmt data-eff
\subfloat[Data efficiency (WMT14)]{
\centering
\resizebox{0.3\textwidth}{!}{
    \begin{tikzpicture}
    \begin{axis}[
        grid=major,
        xlabel=Data Fraction (\%),
        ylabel=BLEU,
        symbolic x coords={25,50,75,100},
        xtick=data,
        xticklabel style={name=tick no \ticknum},
        legend style={at={(0,1)},anchor=north west}
    ]
    \addplot[mark=*,ForestGreen] plot coordinates {
        (25, 25.90)
        (50, 26.20)
        (75, 26.58)
        (100,26.73)
    };
    \addlegendentry{G-XKD}
    \addplot[mark=*,YellowOrange] plot coordinates {
        (25, 25.86)
        (50, 26.15)
        (75, 26.48)
        (100,26.65)
    };
    \addlegendentry{MiniLLM}
    \addplot[mark=*,RoyalBlue] plot coordinates {
        (25, 25.82)
        (50, 26.11)
        (75, 26.50)
        (100,26.61)
    };
    \addlegendentry{GKD}
    \end{axis}
    \end{tikzpicture}
    }
    \label{fig:wmt-data-eff}
}
% gsm data-eff
\subfloat[Data efficiency (GSM8k)]{
\centering
\resizebox{0.3\textwidth}{!}{
    \begin{tikzpicture}
    \begin{axis}[
        grid=major,
        xlabel=Data Fraction (\%),
        ylabel=Accuracy,
        symbolic x coords={25,50,75,100},
        xtick=data,
        xticklabel style={name=tick no \ticknum},
        legend style={at={(0,1)},anchor=north west}
    ]
    \addplot[mark=*,ForestGreen] plot coordinates {
        (25, 7.20)
        (50, 10.45)
        (75, 10.68)
        (100,12.35)
    };
    \addlegendentry{G-XKD}
    \addplot[mark=*,YellowOrange] plot coordinates {
        (25, 7.88)
        (50, 10.00)
        (75, 10.61)
        (100,12.50)
    };
    \addlegendentry{MiniLLM}
    \addplot[mark=*,RoyalBlue] plot coordinates {
        (25, 7.65)
        (50, 10.15)
        (75, 10.53)
        (100,12.35)
    };
    \addlegendentry{GKD}
    \end{axis}
    \end{tikzpicture}
    }
    \label{fig:gsm-data-eff}
}
\caption{Performance-diversity and data efficiency curves of G-XKD, MiniLLM, and GKD.}
\end{figure*}

\subsection{Abstractive Summarization}

Our experiments on the XSum dataset demonstrate that $\mathcal{X}$-KD consistently outperforms the corresponding baseline approaches across both model size configurations.
As shown in Table \ref{tab:overall-results}, under both model size settings, G-XKD outperforms the corresponding baselines GKD and MiniLLM, S-XKD outperforms the corresponding baseline SKD, and SeqXKD outperforms the corresponding baseline SeqKD.

The performance-diversity analysis in Figure \ref{fig:xsum-p-d} reveals that $\mathcal{X}$-KD maintains a better trade-off between summarization quality and diversity than GKD.
Specifically, as diversity increases, the ROUGE-2 score of $\mathcal{X}$-KD decreases more slowly than that of GKD, indicating that experiential learning helps the student model learn more generalizable summarization strategies rather than simply mimicking teacher behavior.

The data efficiency results in Figure \ref{fig:xsum-data-eff} demonstrate that generalized $\mathcal{X}$-KD achieves comparable performance to GKD while using 75\% of the training data, suggesting that incorporating the teacher's reward function helps capture fundamental summarization principles more efficiently.

\subsection{Machine Translation}

On the WMT14 English-to-German translation task, X-KD demonstrates robust improvements across different model sizes.
As shown in Table \ref{tab:overall-results}, under both model size settings, $\mathcal{X}$-KD outperforms all corresponding baseline approaches.

Figure \ref{fig:wmt-p-d} demonstrates the performance-diversity trade-off on the WMT14 English-to-German task.
Unlike the XSum task, not all distillation methods have monotonically decreasing performance-diversity curves on the WMT14 task.
When the sampling temperature is 1.0 (the first point of each curve), the method with higher diversity achieves a higher BLEU score, indicating that assigning appropriate diversity to the decoding process is beneficial to the translation quality.
When lowering the sampling temperature to 0.5 (the second point), the performance of GKD drops significantly, while the performance of experiential learning approaches (i.e., MiniLLM and X-GKD) improves, indicating that experiential learning facilitates the model to jointly achieve translation quality and diversity.
Here, G-XKD achieves a similar BLEU score to MiniLLM, but MiniLLM has better diversity.
When the sampling temperature is 0.3 and 0.1 (the third and fourth points), the performance of the three methods is similar, but G-XKD has the best diversity, indicating that G-XKD can better maintain the translation quality at higher diversity.

The data efficiency analysis in Figure \ref{fig:wmt-data-eff} mirrors the findings from the summarization task, i.e., $\mathcal{X}$-KD requires significantly less training data to achieve performance comparable to baselines trained on the full dataset.
This consistent pattern across tasks suggests that experiential learning helps capture fundamental task principles rather than just surface-level patterns.

\subsection{Arithmetic Reasoning}

For arithmetic reasoning, we evaluate on the GSM8k dataset, which tests the models' ability to solve grade school math word problems.
As shown in Table \ref{tab:overall-results}, in both model settings, SeqXKD and S-XKD outperform SeqKD and SKD, respectively, which indicates that introducing experiential learning into sequence-level and forward-KL distillation improves the prediction accuracy.
However, while G-XKD outperforms all baselines in the large$\rightarrow$base setting, it achieves identical performance to GKD and slightly underperforms MiniLLM in the large$\rightarrow$small setting.
We attribute this to the small capacity of the student model, which results in a bottleneck in arithmetic reasoning ability.

Figure \ref{fig:gsm-p-d} demonstrates the performance-diversity trade-off on the GSM8k task.
At a sampling temperature of 1.0, GKD demonstrates the most balanced performance-diversity characteristics. 
Shifting to temperatures of 0.5 and 0.3, G-XKD emerges as the method with the optimal performance-diversity trade-off.
At a low temperature of 0.1, MiniLLM takes the lead in balancing performance and diversity.
This series of results show that as the sampling temperature decreases and the diversity increases, experiential learning, especially the MiniLLM training objective, facilitates the model to maintain better arithmetic reasoning accuracies.
However, at the temperature of 0.3, all three approaches achieve their best performances that are similar to each other, while G-XKD achieves the highest diversity.
This observation suggests that G-XKD can achieve a good performance-diversity trade-off in arithmetic reasoning if the decoding temperature is tuned optimal.

The data efficiency results in Figure \ref{fig:gsm-data-eff} show that when using 25\% of the training data, G-XKD underperforms MiniLLM and GKD.
However, when using 50\% of the training data, G-XKD outperforms them and is even comparable to the performance achieved by these two approaches using 75\% of the training data.
When the training data ratio is 75\% and 100\%, the three methods achieve similar performance.

\subsection{Black-Box Distillation}

We perform black-box distillation using SeqKD and SeqXKD, where Llama-2-7b is adopted as the teacher model and T5-large is adopted as the student model.
We take the first 1000 data in each dataset and use Llama-2-7b to generate 10 answers for each source sentence in the dataset, thereby constructing an offline teacher behavior dataset $\mathcal{D}_\text{teacher}$.
Then, in line 7 of Algorithm \ref{alg:blackbox-distill}, instead of sampling the response from the online teacher model $\pi^*$, we directly use the response corresponding to $\mathbf{x}$ in $\mathcal{D}_\text{teacher}$.
The distillation of black-box models does not use the ground truth in the dataset as supervision.
Therefore, instead of evaluating the automatic metrics (ROUGE-2, BLEU, Accuracy), we employ GPT-4 to evaluate which response is better and report the win rate of our method SeqXKD against the baseline SeqKD.

\begin{table}[]
\centering
\caption{GPT-4 win rate evaluation of SeqXKD against SeqKD.}
\label{tab:black-box-wtl}
\begin{tabular}{@{}lrrr@{}}
\toprule
\multicolumn{1}{c}{} & \multicolumn{1}{r}{Win} & \multicolumn{1}{r}{Tie} & \multicolumn{1}{r}{Lose} \\ \midrule
XSum                 & 48.3\%                  & 8.7\%                   & 43.0\%                    \\
WMT                  & 43.8\%                  & 13.6\%                  & 42.6\%                    \\
GSM                  & 48.9\%                  & 2.8\%                   & 48.3\%                    \\ \bottomrule
\end{tabular}
\end{table}

Results in Table \ref{tab:black-box-wtl} show that SeqXKD outperforms SeqKD in win rate evaluation.
In addition, the SeqXKD and SeqKD experiments in Table \ref{tab:overall-results} are also black-box KD using T5-large as the black-box teacher model, where the results of SeqXKD are also consistently higher than SeqKD.
These findings indicate that SeqXKD with experiential learning provides better performance in black-box distillation.

\section{Conclusion}

We present $\mathcal{X}$-KD, a novel knowledge distillation framework for LLMs.
By leveraging Bayesian inverse reinforcement learning, the framework enables student models to capture the underlying reward function of the teacher's learning environment.
$\mathcal{X}$-KD's flexibility in integrating with different KD techniques and simplicity in maintaining a supervised learning paradigm highlight its potential for various applications.
Experimental results show that $\mathcal{X}$-KD outperforms existing KD methods by delivering better performance, balancing performance and diversity, and improving data efficiency.
However, we have to acknowledge that this paper has some limitations.
Firstly, $\mathcal{X}$-KD's performance is sensitive to the tuning of the experiential weight, which increases the cost of hyperparameter tuning in practice.
Secondly, the paper mainly focuses on the white-box setting and pays less attention to the black-box setting claimed applicable.
We will conduct a more systematic study with $\mathcal{X}$-KD in black-box distillation in the future.

\section*{Impact Statement}

This paper presents work whose goal is to advance the field of Machine Learning. There are many potential societal consequences of our work, none which we feel must be specifically highlighted here.

\bibliography{example_paper}

@book{kolb2014experiential,
  title={Experiential learning: Experience as the source of learning and development},
  author={Kolb, David A},
  year={2014},
  publisher={FT press}
}

@inproceedings{ng2000algorithms,
  title={Algorithms for inverse reinforcement learning.},
  author={Ng, Andrew Y and Russell, Stuart and others},
  booktitle={Icml},
  volume={1},
  pages={2},
  year={2000}
}

@inproceedings{bain1995framework,
  title={A Framework for Behavioural Cloning.},
  author={Bain, Michael and Sammut, Claude},
  booktitle={Machine Intelligence 15},
  pages={103--129},
  year={1995}
}

@article{brown2020language,
  title={Language models are few-shot learners},
  author={Brown, Tom and Mann, Benjamin and Ryder, Nick and Subbiah, Melanie and Kaplan, Jared D and Dhariwal, Prafulla and Neelakantan, Arvind and Shyam, Pranav and Sastry, Girish and Askell, Amanda and others},
  journal={Advances in neural information processing systems},
  volume={33},
  pages={1877--1901},
  year={2020}
}

@article{raffel2020exploring,
  title={Exploring the limits of transfer learning with a unified text-to-text transformer},
  author={Raffel, Colin and Shazeer, Noam and Roberts, Adam and Lee, Katherine and Narang, Sharan and Matena, Michael and Zhou, Yanqi and Li, Wei and Liu, Peter J},
  journal={Journal of machine learning research},
  volume={21},
  number={140},
  pages={1--67},
  year={2020}
}

@article{openai2022chat,
  title={Introducing ChatGPT},
  author={OpenAI},
  journal={https://openai.com/index/chatgpt/},
  year={2022}
}

@article{achiam2023gpt,
  title={Gpt-4 technical report},
  author={Achiam, Josh and Adler, Steven and Agarwal, Sandhini and Ahmad, Lama and Akkaya, Ilge and Aleman, Florencia Leoni and Almeida, Diogo and Altenschmidt, Janko and Altman, Sam and Anadkat, Shyamal and others},
  journal={arXiv preprint arXiv:2303.08774},
  year={2023}
}

@article{kim2016sequence,
  title={Sequence-level knowledge distillation},
  author={Kim, Yoon and Rush, Alexander M},
  journal={arXiv preprint arXiv:1606.07947},
  year={2016}
}

@article{sanh2019distilbert,
  title={DistilBERT, a distilled version of BERT: smaller, faster, cheaper and lighter},
  author={Sanh, Victor and Debut, Lysandre and Chaumond, Julien and Wolf, Thomas},
  journal={arXiv preprint arXiv:1910.01108},
  year={2019}
}

@article{wen2023f,
  title={f-Divergence Minimization for Sequence-Level Knowledge Distillation},
  author={Wen, Yuqiao and Li, Zichao and Du, Wenyu and Mou, Lili},
  journal={arXiv preprint arXiv:2307.15190},
  year={2023}
}

@inproceedings{liang2023less,
  title={Less is more: Task-aware layer-wise distillation for language model compression},
  author={Liang, Chen and Zuo, Simiao and Zhang, Qingru and He, Pengcheng and Chen, Weizhu and Zhao, Tuo},
  booktitle={International Conference on Machine Learning},
  pages={20852--20867},
  year={2023},
  organization={PMLR}
}

@inproceedings{gu2024minillm,
  title={MiniLLM: Knowledge distillation of large language models},
  author={Gu, Yuxian and Dong, Li and Wei, Furu and Huang, Minlie},
  booktitle={The Twelfth International Conference on Learning Representations},
  year={2024}
}

@inproceedings{agarwal2024policy,
  title={On-policy distillation of language models: Learning from self-generated mistakes},
  author={Agarwal, Rishabh and Vieillard, Nino and Zhou, Yongchao and Stanczyk, Piotr and Garea, Sabela Ramos and Geist, Matthieu and Bachem, Olivier},
  booktitle={The Twelfth International Conference on Learning Representations},
  year={2024}
}

@article{xu2024survey,
  title={A survey on knowledge distillation of large language models},
  author={Xu, Xiaohan and Li, Ming and Tao, Chongyang and Shen, Tao and Cheng, Reynold and Li, Jinyang and Xu, Can and Tao, Dacheng and Zhou, Tianyi},
  journal={arXiv preprint arXiv:2402.13116},
  year={2024}
}

@inproceedings{chan2021scalable,
    title={Scalable {B}ayesian Inverse Reinforcement Learning},
    author={Alex James Chan and Mihaela van der Schaar},
    booktitle={International Conference on Learning Representations},
    year={2021},
    url={https://openreview.net/forum?id=4qR3coiNaIv}
}

@inproceedings{ranzato2016sequence,
  title={Sequence level training with recurrent neural networks},
  author={Ranzato, Marc’Aurelio and Chopra, Sumit and Auli, Michael and Zaremba, Wojciech},
  booktitle={4th International Conference on Learning Representations, ICLR 2016},
  year={2016}
}

@article{hinton2015distilling,
  title={Distilling the knowledge in a neural network},
  author={Hinton, Geoffrey and Vinyals, Oriol and Dean, Jeff},
  journal={arXiv preprint arXiv:1503.02531},
  year={2015}
}

@article{cobbe2021training,
  title={Training verifiers to solve math word problems},
  author={Cobbe, Karl and Kosaraju, Vineet and Bavarian, Mohammad and Chen, Mark and Jun, Heewoo and Kaiser, Lukasz and Plappert, Matthias and Tworek, Jerry and Hilton, Jacob and Nakano, Reiichiro and others},
  journal={arXiv preprint arXiv:2110.14168},
  year={2021}
}

@article{narayan2018don,
  title={Don't give me the details, just the summary! topic-aware convolutional neural networks for extreme summarization},
  author={Narayan, Shashi and Cohen, Shay B and Lapata, Mirella},
  journal={arXiv preprint arXiv:1808.08745},
  year={2018}
}

@inproceedings{bojar2014findings,
  title={Findings of the 2014 workshop on statistical machine translation},
  author={Bojar, Ond{\v{r}}ej and Buck, Christian and Federmann, Christian and Haddow, Barry and Koehn, Philipp and Leveling, Johannes and Monz, Christof and Pecina, Pavel and Post, Matt and Saint-Amand, Herve and others},
  booktitle={Proceedings of the ninth workshop on statistical machine translation},
  pages={12--58},
  year={2014}
}

@inproceedings{yim2017gift,
  title={A gift from knowledge distillation: Fast optimization, network minimization and transfer learning},
  author={Yim, Junho and Joo, Donggyu and Bae, Jihoon and Kim, Junmo},
  booktitle={Proceedings of the IEEE conference on computer vision and pattern recognition},
  pages={4133--4141},
  year={2017}
}

@article{dettmers2023qlora,
  title={QLoRA: efficient finetuning of quantized LLMs (2023)},
  author={Dettmers, Tim and Pagnoni, Artidoro and Holtzman, Ari and Zettlemoyer, Luke},
  journal={arXiv preprint arXiv:2305.14314},
  volume={52},
  pages={3982--3992},
  year={2023}
}

@article{hu2021lora,
  title={Lora: Low-rank adaptation of large language models},
  author={Hu, Edward J and Shen, Yelong and Wallis, Phillip and Allen-Zhu, Zeyuan and Li, Yuanzhi and Wang, Shean and Wang, Lu and Chen, Weizhu},
  journal={arXiv preprint arXiv:2106.09685},
  year={2021}
}

@book{sutton2018reinforcement,
  title={Reinforcement learning: An introduction},
  author={Sutton, Richard S and Barto, Andrew G},
  year={2018},
  publisher={MIT press}
}
\bibliographystyle{icml2025}

%%%%%%%%%%%%%%%%%%%%%%%%%%%%%%%%%%%%%%%%%%%%%%%%%%%%%%%%%%%%%%%%%%%%%%%%%%%%%%%
%%%%%%%%%%%%%%%%%%%%%%%%%%%%%%%%%%%%%%%%%%%%%%%%%%%%%%%%%%%%%%%%%%%%%%%%%%%%%%%
% APPENDIX
%%%%%%%%%%%%%%%%%%%%%%%%%%%%%%%%%%%%%%%%%%%%%%%%%%%%%%%%%%%%%%%%%%%%%%%%%%%%%%%
%%%%%%%%%%%%%%%%%%%%%%%%%%%%%%%%%%%%%%%%%%%%%%%%%%%%%%%%%%%%%%%%%%%%%%%%%%%%%%%
\newpage
\appendix

\section{Derivations}
\label{appendix:derive}

\subsection{Original Reward Modeling}
\label{appendix:derive-orm}

The original reward modeling objective is to learn the reward function based on which the teacher policy $\pi$ is trained.
In Section \ref{sec:orm}, we intuitively apply the AVRIL framework to the original reward modeling ELBO objective in Equation \ref{eq:org_reward_elbo} and obtain the training objective in Equation \ref{eq:org_reward_avril}.
Here, we give the detailed derivation.

Given the ELBO objective:
\begin{equation}
\label{eq:dev-1-1}
\max_\phi\mathbb{E}_{R\sim q_\phi(\cdot)}[\log p(\pi|R)]-D_\text{KL}[q_\phi(R)||p(R)]
\end{equation}
The likelihood $p(\pi|R)$ denotes the probability of obtaining policy $\pi$ under the reward function $R$.
Let $\pi_R$ be the optimal policy under reward $R$, $\mathcal{D}_\pi\sim\pi(\cdot)$ is the behavioral data sampled from policy $\pi$.
Then, $p(\pi|R)$ can be regarded as $p(\mathcal{D}_\pi|\pi_R)$, which means the probability of behavior $\mathcal{D}_\pi$ under policy $\pi_R$ and can be further denoted as Equation \ref{eq:dev-1-2}, where $\pi_R(a|s)$ is the probability of action $a$ given state $s$ under policy $\pi_R$.
\begin{equation}
\label{eq:dev-1-2}
p(\pi|R)=p(\mathcal{D}_\pi|\pi_R)=\mathbb{E}_{(s,a)\sim\pi}[\pi_R(a|s)]
\end{equation}
The policy $\pi_R$ can be formulated as a Boltzmann policy, as shown in Equation \ref{eq:dev-1-3}, where $Q^{\pi_R}_R$ is the state-action value function under reward function $R$ and policy $\pi_R$.
\begin{equation}
\label{eq:dev-1-3}
\pi_R(a|s)=\frac{\exp(\tau Q^{\pi_R}_R(s,a))}{\sum_{b\in\mathcal{A}}\exp(\tau Q^{\pi_R}_R(s,b))}
\end{equation}
By substituting Equation \ref{eq:dev-1-3} into the ELBO objective, we obtain an equivalent objective, as shown in Equation \ref{eq:dev-1-3-1}.
Note that here we move the $\log$ into the expectation over $(s,a)\sim\pi$, which does not affect the optimality of the training objective.
\begin{equation}
\label{eq:dev-1-3-1}
\max_{\phi}\mathbb{E}_{R\sim q_\phi}\left[
\begin{aligned}
\log \mathbb{E}_{(s,a)\sim\pi}\left[\frac{\exp(\tau Q^{\pi_R}_R(s,a))}{\sum_{b\in\mathcal{A}}\exp(\tau Q^{\pi_R}_R(s,b))}\right] \\
-D_\text{KL}[q_\phi(R)||p(R)]
\end{aligned}
\right]
\end{equation}

To evaluate $Q^{\pi_R}_R(s,a)$, we use a neural network $Q_\theta$ parameterized by $\theta$ to approximate the state-action value:
\begin{equation}
\label{eq:dev-1-4}
Q^{\pi_R}_R(s,a)\approx Q_\theta(s,a)
\end{equation}
By directly replacing $Q^{\pi_R}_R$ with $Q_\theta(s,a)$ in Equation \ref{eq:dev-1-3-1} and considering the optimization of the newly introduced parameter $\theta$, we obtain a new training objective:
\begin{equation}
\label{eq:dev-1-6}
\max_{\phi,\theta}\mathbb{E}_{R\sim q_\phi}\left[
\begin{aligned}
\log \mathbb{E}_{(s,a)\sim\pi}\left[\frac{\exp(\tau Q_\theta(s,a))}{\sum_{b\in\mathcal{A}}\exp(\tau Q_\theta(s,b))}\right] \\
-D_\text{KL}[q_\phi(R)||p(R)]
\end{aligned}
\right]
\end{equation}
However, this objective is not equivalent to the original ELBO objective in Equation \ref{eq:dev-1-1}.
The original objective (Equation \ref{eq:dev-1-3-1}) maximizes the parameter $\phi$ based on the strict state-action value function $Q^{\pi_R}_R$, while Equation \ref{eq:dev-1-6} maximizes the approximated state-action value function $Q_\theta$.
This sounds feasible, but there is no objective to draw $Q_\theta$ close to $Q^{\pi_R}_R$.
Therefore, $Q_\theta$ may become completely different from $Q^{\pi_R}_R$, and thus the objective in Equation \ref{eq:dev-1-6} is completely different from that in Equation \ref{eq:dev-1-3-1}.

To address the problem, the AVRIL framework introduces a constraint term.
According to the Bellman equation \cite{sutton2018reinforcement}, the relationship between reward function $R$ and the state-action value function $Q^{\pi_R}_R$ under this reward satisfies the following equation:
\begin{equation}
\label{eq:dev-1-7}
R(s,a)=\mathbb{E}_{(s',a')\sim\pi_R}[Q^{\pi_R}_R(s,a)-\gamma Q^{\pi_R}_R(s',a')]
\end{equation}
In other words, for the parameterized state-action value function $Q_\theta$, if $Q_\theta$ is trained under the reward function $R$, then $Q_\theta$ must satisfy Equation \ref{eq:dev-1-8} for any $(s,a)$.
\begin{equation}
\label{eq:dev-1-8}
\mathbb{E}_{(s',a')\sim\pi_R}[Q_\theta(s,a)-\gamma Q_\theta(s',a')]=R(s,a)
\end{equation}
Therefore, AVRIL adds a constraint to the training objective in Equation \ref{eq:dev-1-6} and obtains the training objective in Equation \ref{eq:dev-1-9}.
\begin{equation}
\label{eq:dev-1-9}
\begin{aligned}
&\max_{\phi,\theta}\mathbb{E}_{R\sim q_\phi}\left[
\begin{aligned}
\mathbb{E}_{(s,a)\sim\pi}\left[\log\frac{\exp(\tau Q_\theta(s,a))}{\sum_{b\in\mathcal{A}}\exp(\tau Q_\theta(s,b))}\right] \\
-D_\text{KL}[q_\phi(R)||p(R)]
\end{aligned}
\right] \\
&\text{subject to} \\
&\mathbb{E}_{(s,a,s',a')\sim\pi}[-\log q_\phi(Q_\theta(s,a)-\gamma Q_\theta(s',a'))]<\epsilon
\end{aligned}
\end{equation}
Rewriting Equation \ref{eq:dev-1-9} as a Lagrangian under the KKT conditions, and given complimentary slackness, a practical objective function can be obtained, as shown in Equation \ref{eq:dev-1-10}.
Note that here we remove the expectation over $R\sim q_\phi$ for approximation, which is also performed in AVRIL \cite{chan2021scalable}.
Equation \ref{eq:dev-1-10} is exactly the ORM training objective in Equation \ref{eq:org_reward_avril}.
\begin{equation}
\label{eq:dev-1-10}
\max_{\phi,\theta}
\mathbb{E}_{(s,a,s',a')\sim\pi}\left[
\begin{aligned}
\frac{\log\exp(\tau Q_\theta(s,a))}{\sum_{b\in\mathcal{A}}\exp(\tau Q_\theta(s,b))} \\
-D_\text{KL}[q_\phi(\cdot|s,a)||p(\cdot)] \\
+\lambda\log q_\phi(Q_\theta(s,a)-\gamma Q_\theta(s',a'))
\end{aligned}
\right]
\end{equation}

\subsection{ORM in NLG Setting}
\label{appendix:derive-orm-nlg}

Determining the meaning of the state and action quadruple $(s,a,s',s')$ in Equation \ref{eq:org_reward_avril} (i.e., \ref{eq:dev-1-10}) is the first step in applying the ORM training objective to a specific problem.
As clarified in Section \ref{sec:orm}, in the NLG setting, given the prompt $\mathbf{x}$, at timestep $t$, the current state is $s=(\mathbf{x},\mathbf{y}_{1:t})$, the current action is $a=y_{t+1}$, the next state is $s'=(\mathbf{x},\mathbf{y}_{1:t+1})$, and the next action is $a'=y_{t+2}$.
Therefore, the sampling of $a$ from policy $\pi$ given the state $s$, which is originally denoted as $a\sim\pi(\cdot|s)$, is now denoted as $y_{t+1}\sim\pi(\cdot|\mathbf{x},\mathbf{y}_{1:t})$, which is the auto-regressive generation given $\pi$ as the language model.
Furthermore, the expectation over state-action quadruples $\mathbb{E}_{(s,a,s',a')\sim\pi}[f(s,a,s',a')]$ is now denoted as:
\begin{equation}
\label{eq:dev-2-1}
\mathbb{E}_{\mathbf{x}\sim\mathcal{D}}\mathbb{E}_{\mathbf{y}\sim\pi(\cdot|\mathbf{x})}\sum_{t=1}^{|\mathbf{y}|}f\big(\overbrace{(\mathbf{x},\mathbf{y}_{1:t})}^{s},\underbrace{y_{t+1}}_{a},\overbrace{(\mathbf{x},\mathbf{y}_{1:t+1})}^{s'},\underbrace{y_{t+2}}_{a'}\big)
\end{equation}
where $\mathcal{D}$ is the prompt dataset denoting the initial state distribution.

Applying the format in Equation \ref{eq:dev-2-1} to the ORM training objective in Equation \ref{eq:dev-1-10}, we obtain the ORM training objective in the NLG setting, as shown in Equation \ref{eq:dev-2-2}, which is exactly the objective in Equation \ref{eq:org_reward_nlg}.
\begin{equation}
\label{eq:dev-2-2}
\begin{aligned}
&-\sum_{\mathbf{x}\in\mathcal{D}} \\
&\mathbb{E}_{\mathbf{y}\sim\pi(\cdot|\mathbf{x})}\left[
\sum_{t=1}^{|\mathbf{y}|}\left[
\begin{aligned}
\log \frac{\exp(\tau Q_\theta((\mathbf{x},\mathbf{y}_{1:t}),y_{t+1}))}{\sum_{y'\in\mathcal{V}}\exp(\tau Q_\theta((\mathbf{x},\mathbf{y}_{1:t}),y'))} \\
-D_\text{KL}[q_\phi(\cdot|\mathbf{x},\mathbf{y}_{1:t+1})||p(\cdot)] \\
+\lambda \log q_\phi(\delta_t(\theta)|\mathbf{x},\mathbf{y}_{1:t+1})
\end{aligned}
\right]
\right]
\end{aligned}
\end{equation}

\subsection{Sequence-Level KD Loss Reformulation}
\label{appendix:derive-seqkd}

In Section \ref{sec:s-xkd}, we reformulate the sequence-level loss function as Equation \ref{eq:seq-reform}, which separates the forward-KL $D_\text{KL}(\pi(\cdot|\mathbf{x})||B_\theta(\cdot|\mathbf{x}))$ from the sequence-level loss.
The detailed derivation is as follows:
\begin{equation*}
\begin{aligned}
\mathcal{L}_\text{seq}&(\theta)=-\sum_{\mathbf{x}\in\mathcal{D}}\mathbb{E}_{\mathbf{y}\sim\pi(\cdot|\mathbf{x})}[\log B_\theta(\mathbf{y}|\mathbf{x})] \\
&=-\sum_{\mathbf{x}\in\mathcal{D}}\mathbb{E}_{\mathbf{y}\sim\pi(\cdot|\mathbf{x})}\left[
\begin{aligned}
\log B_\theta(\mathbf{y}|\mathbf{x})-\log \pi(\mathbf{y}|\mathbf{x}) \\
+\log \pi(\mathbf{y}|\mathbf{x})
\end{aligned}
\right] \\
&=-\sum_{\mathbf{x}\in\mathcal{D}}\mathbb{E}_{\mathbf{y}\sim\pi(\cdot|\mathbf{x})}\left[\log \frac{B_\theta(\mathbf{y}|\mathbf{x})}{\pi(\mathbf{y}|\mathbf{x})}+\log \pi(\mathbf{y}|\mathbf{x})\right] \\
&=-\sum_{\mathbf{x}\in\mathcal{D}}\mathbb{E}_{\mathbf{y}\sim\pi(\cdot|\mathbf{x})}\left[-\log \frac{\pi(\mathbf{y}|\mathbf{x})}{B_\theta(\mathbf{y}|\mathbf{x})}+\log \pi(\mathbf{y}|\mathbf{x})\right] \\
&=\sum_{\mathbf{x}\in\mathcal{D}}\mathbb{E}_{\mathbf{y}\sim\pi(\cdot|\mathbf{x})}\left[\log \frac{\pi(\mathbf{y}|\mathbf{x})}{B_\theta(\mathbf{y}|\mathbf{x})}-\log \pi(\mathbf{y}|\mathbf{x})\right] \\
&=\sum_{\mathbf{x}\in\mathcal{D}}
\left[
\begin{aligned}
D_\text{KL}[\pi(\cdot|\mathbf{x})||B_\theta(\cdot|\mathbf{x})] \\
\color{red}{-\mathbb{E}_{\mathbf{y}\sim\pi(\cdot|\mathbf{x})}[\log \pi(\mathbf{y}|\mathbf{x})]}
\end{aligned}
\right]
\end{aligned}
\end{equation*}

In Section \ref{sec:g-xkd}, we reformulate the combination of forward and reverse sequence-level loss functions (Equation \ref{eq:seqkd-gen}) as Equation \ref{eq:gen-seq-reform}, which separates the Jensen-Shannon divergence $D_\beta(\pi(\cdot|\mathbf{x})||B_\theta(\cdot|\mathbf{x}))$ from the sequence-level loss.
The detailed derivation is as follows:
\begin{equation*}
\begin{aligned}
\mathcal{L}_\text{g-seq}&(\theta)=\beta\cdot\mathcal{L}_\text{seq}(\theta)+(1-\beta)\cdot\mathcal{L}_\text{r-seq}(\theta) \\
&=-\beta\cdot\sum_{\mathbf{x}\in\mathcal{D}}\mathbb{E}_{\mathbf{y}\sim\pi(\cdot|\mathbf{x})}[\log B_\theta(\mathbf{y}|\mathbf{x})] \\
&~~~~-(1-\beta)\cdot\sum_{\mathbf{x}\in\mathcal{D}}\mathbb{E}_{\mathbf{y}\sim B_\theta(\cdot|\mathbf{x})}[\log \pi(\mathbf{y}|\mathbf{x})] \\
&=\beta\cdot\sum_{\mathbf{x}\in\mathcal{D}}
\left[
\begin{aligned}
D_\text{KL}[\pi(\cdot|\mathbf{x})||B_\theta(\cdot|\mathbf{x})] \\
-\mathbb{E}_{\mathbf{y}\sim\pi(\cdot|\mathbf{x})}[\log \pi(\mathbf{y}|\mathbf{x})]
\end{aligned}
\right] \\
&~~~~-(1-\beta)\cdot\sum_{\mathbf{x}\in\mathcal{D}}\mathbb{E}_{\mathbf{y}\sim B_\theta(\cdot|\mathbf{x})}
\left[
\begin{aligned}
\log \frac{\pi(\mathbf{y}|\mathbf{x})}{B_\theta(\mathbf{y}|\mathbf{x})} \\
+\log B_\theta(\mathbf{y}|\mathbf{x})
\end{aligned}
\right] \\
&=\beta\cdot\sum_{\mathbf{x}\in\mathcal{D}}
\left[
\begin{aligned}
D_\text{KL}[\pi(\cdot|\mathbf{x})||B_\theta(\cdot|\mathbf{x})] \\
-\mathbb{E}_{\mathbf{y}\sim\pi(\cdot|\mathbf{x})}[\log \pi(\mathbf{y}|\mathbf{x})]
\end{aligned}
\right] \\
&~~~~+(1-\beta)\cdot\sum_{\mathbf{x}\in\mathcal{D}}
\left[
\begin{aligned}
D_\text{KL}[B_\theta(\cdot|\mathbf{x})||\pi(\cdot|\mathbf{x})] \\
-\mathbb{E}_{\mathbf{y}\sim B_\theta(\cdot|\mathbf{x})}[\log B_\theta(\mathbf{y}|\mathbf{x})]
\end{aligned}
\right] \\
&=\sum_{\mathbf{x}\in\mathcal{D}}
\left[
\begin{aligned}
\beta\cdot D_\text{KL}[\pi(\cdot|\mathbf{x})||B_\theta(\cdot|\mathbf{x})] \\
+(1-\beta)\cdot D_\text{KL}[B_\theta(\cdot|\mathbf{x})||\pi(\cdot|\mathbf{x})] \\
-\beta\mathbb{E}_{\mathbf{y}\sim\pi(\cdot|\mathbf{x})}[\log \pi(\mathbf{y}|\mathbf{x})] \\
-(1-\beta)\mathbb{E}_{\mathbf{y}\sim B_\theta(\cdot|\mathbf{x})}[\log B_\theta(\mathbf{y}|\mathbf{x})]
\end{aligned}
\right] \\
&=\sum_{\mathbf{x}\in\mathcal{D}}
\left[
\begin{aligned}
D_\beta[\pi(\cdot|\mathbf{x})||B_\theta(\cdot|\mathbf{x})] \\
\color{red}{-\beta\mathbb{E}_{\mathbf{y}\sim\pi(\cdot|\mathbf{x})}[\log \pi(\mathbf{y}|\mathbf{x})]} \\
\color{red}{-(1-\beta)\mathbb{E}_{\mathbf{y}\sim B_\theta(\cdot|\mathbf{x})}[\log B_\theta(\mathbf{y}|\mathbf{x})]}
\end{aligned}
\right]
\end{aligned}
\end{equation*}

The red terms above are ignored during optimization.
Specifically, $\color{red}{\mathbb{E}_{\mathbf{y}\sim\pi(\cdot|\mathbf{x})}[\log\pi(\mathbf{y}|\mathbf{x})]}$ is not relevant to the parameter $\theta$ and is therefore ignored in optimization.
On the other hand, maximizing $\color{red}{\mathbb{E}_{\mathbf{y}\sim B_\theta(\cdot|\mathbf{x})}[\log B_\theta(\mathbf{y}|\mathbf{x})]}$ literally means maximizing the probability of sentences generated by the $B_\theta$ strategy under $B_\theta$, which is redundant and therefore also ignored.
Therefore, minimizing $\mathcal{L}_\text{seq}(\theta)$ is equivalent to minimizing the forward KL divergence between $\pi$ and $B_\theta$, and minimizing $\mathcal{L}_\text{g-seq}(\theta)$ is equivalent to minimizing the Jesen-Shannon divergence between $\pi$ and $B_\theta$.

\section{Additional Experiments}
\label{appendix:addexp}

\subsection{Scalability Study}

\begin{figure}[H]
\centering
\resizebox{0.4\textwidth}{!}{
\begin{tikzpicture}
\begin{axis}[
    grid=major,
    ylabel=ROUGE-2,
    xticklabels={base$\rightarrow$small,large$\rightarrow$small,xl$\rightarrow$small,xl$\rightarrow$base,xl$\rightarrow$large},
    xtick={1,2,3,4,5},
    xticklabel style={name=tick no \ticknum,rotate=60},
    legend style={at={(0,1)},anchor=north west}
]
\addplot[mark=*,ForestGreen] plot coordinates {
    (1,13.22)
    (2,13.81)
    (3,16.42)
    (4,18.55)
    (5,21.14)
};
\addlegendentry{G-XKD}
\addplot[mark=*,RoyalBlue] plot coordinates {
    (1,13.04)
    (2,13.62)
    (3,16.15)
    (4,18.38)
    (5,20.84)
};
\addlegendentry{GKD}
\end{axis}
\end{tikzpicture}
}
\caption{Scalability of G-XKD and GKD.}
\label{fig:scalability}
\end{figure}
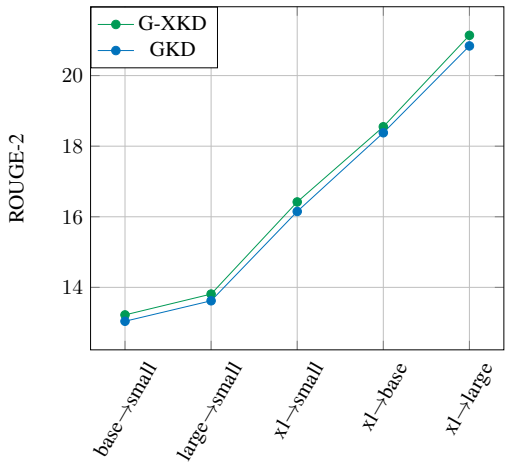

To investigate the scalability of $\mathcal{X}$-KD, we examine how variations in model size, for both teacher and student architectures, influence distillation performance.
While our primary analysis employs T5-small and T5-base as student models with T5-large as the teacher model, we conduct additional experiments across more detailed configurations.
In one set of experiments, we maintain T5-small as the student model while systematically employing larger teacher models: T5-base, T5-large, and T5-xl.
Conversely, we fix T5-xl as the teacher model and evaluate the performance using progressively larger student models: T5-small, T5-base, and T5-large.
We implement GKD and G-XKD distillation approaches on the XSum dataset across these configurations, with results presented in Figure \ref{fig:scalability}.
The empirical evidence demonstrates that G-XKD consistently achieves superior performance compared to GKD, regardless of the size increases in either the teacher or student model.

\subsection{Self-Distillation}

Self-distillation \cite{yim2017gift} involves distilling knowledge from a teacher model to a student model with the same size and architecture.
GKD demonstrates its superior performance on self-distillation \cite{agarwal2024policy}.
We seek to evaluate whether $\mathcal{X}$-KD could surpass the performance of non-experiential KD approaches in self-distillation.
To this end, we conducted comprehensive self-distillation experiments across the previously discussed three NLG tasks, implementing both G-XKD and GKD training objectives using the T5-large architecture.
As illustrated in Table \ref{tab:self-distill}, G-XKD consistently demonstrates superior performance across all evaluation tasks.
This consistent outperformance suggests that the experiential learning paradigm inherent in G-XKD offers substantial advantages even in scenarios where architectural differences between teacher and student models are eliminated.

\begin{table}[]
\centering
\caption{Self-distillation performances on three tasks.}
\label{tab:self-distill}
\begin{tabular}{@{}llll@{}}
\toprule
      & XSum  & WMT   & GSM8k \\ \midrule
GKD   & 18.33 & 27.48 & 23.33 \\
G-XKD & \textbf{18.46} & \textbf{27.67} & \textbf{23.48} \\ \bottomrule
\end{tabular}
\end{table}

\subsection{Experiential Weight Ablation}

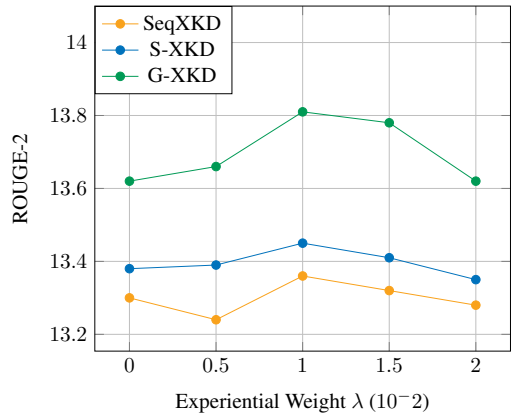
\begin{figure}[H]
\centering
\resizebox{0.4\textwidth}{!}{
\begin{tikzpicture}
\begin{axis}[
    grid=major,
    ymax=14.1,
    xlabel=Experiential Weight $\lambda$ ($10^-2$),
    ylabel=ROUGE-2,
    xtick={0.0,0.5,1.0,1.5,2.0},
    legend style={at={(0,1)},anchor=north west}
]
\addplot[mark=*,YellowOrange] plot coordinates {
    (0.0,13.30)
    (0.5,13.24)
    (1.0,13.36)
    (1.5,13.32)
    (2.0,13.28)
};
\addlegendentry{SeqXKD}
\addplot[mark=*,RoyalBlue] plot coordinates {
    (0.0,13.38)
    (0.5,13.39)
    (1.0,13.45)
    (1.5,13.41)
    (2.0,13.35)
};
\addlegendentry{S-XKD}
\addplot[mark=*,ForestGreen] plot coordinates {
    (0.0,13.62)
    (0.5,13.66)
    (1.0,13.81)
    (1.5,13.78)
    (2.0,13.62)
};
\addlegendentry{G-XKD}
\end{axis}
\end{tikzpicture}
}
\caption{Performance under different experiential weights.}
\label{fig:lambda-ablation}
\end{figure}

The experiential weight $\lambda$ in Equation \ref{eq:fex} controls the degree of experiential learning since it scales the consistency between the policy and the reward function.
In our primary experimental setup, we consistently maintained $\lambda$ at 0.001 across all $\mathcal{X}$-KD methods.
To comprehensively analyze the impact of the experiential weight, we conducted a systematic investigation by progressively increasing $\lambda$ from 0.000 to 0.002, enabling us to observe the consequent effects on $\mathcal{X}$-KD performance metrics.
Specifically, we implement SeqXD, S-XKD, and G-XKD under T5-large$\rightarrow$T5-small setting with different $\lambda$ configurations on the XSum dataset.
As illustrated in Figure \ref{fig:lambda-ablation}, as $\lambda$ increases, the performance of all KD approaches shows a trend of first increasing and then decreasing, achieving the highest performance at $\lambda=0.001$.
This empirical observation underscores a notable limitation in our methodology, specifically its sensitivity to hyperparameter modifications.
This susceptibility to parameter variation indicates that our training objective would benefit from additional normalization techniques.

\subsection{Boltzmann Temperature Ablation}

\begin{figure}[H]
\centering
\resizebox{0.4\textwidth}{!}{
\begin{tikzpicture}
\begin{axis}[
    ymax=14.1,
    xlabel=Boltzmann temperature $\tau'$,
    ylabel=ROUGE-2,
    xtick={0.1,0.3,0.5,1.0},
    legend style={at={(0,1)},anchor=north west}
]
\addplot[mark=*,YellowOrange] plot coordinates {
    (0.1,13.33)
    (0.3,13.34)
    (0.5,13.36)
    (1.0,13.36)
};
\addlegendentry{SeqXKD}
\addplot[mark=*,RoyalBlue] plot coordinates {
    (0.1,13.42)
    (0.3,13.43)
    (0.5,13.43)
    (1.0,13.45)
};
\addlegendentry{S-XKD}
\addplot[mark=*,ForestGreen] plot coordinates {
    (0.1,13.74)
    (0.3,13.76)
    (0.5,13.79)
    (1.0,13.81)
};
\addlegendentry{G-XKD}
\draw[dashed,thick,YellowOrange] (axis cs:0,13.30) -- (axis cs:1.1,13.30);
\draw[dashed,thick,RoyalBlue] (axis cs:0,13.38) -- (axis cs:1.1,13.38);
\draw[dashed,thick,ForestGreen] (axis cs:0,13.62) -- (axis cs:1.1,13.62);
\end{axis}
\end{tikzpicture}
}
\caption{Performance under different Boltzmann temperature configurations. The dashed lines of different colors represent the performance of the corresponding non-experiential baselines.}
\label{fig:tau-ablation}
\end{figure}
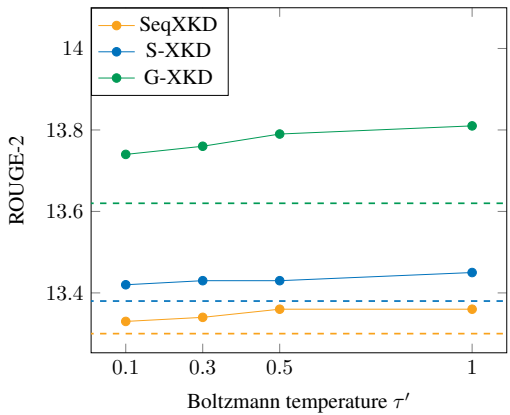

The Boltzmann temperature $\tau'$ in Equation \ref{eq:inverse-boltz} controls the correlation between the Q value of each action and its probability under the LLM policy.
A higher Boltzmann temperature indicates that the Q value is more affected by the action probability.
In our primary experimental setup, we set $\tau'$ to 1.0 across all $\mathcal{X}$-KD methods.
To evaluate the sensitivity of our framework to this crucial hyperparameter, we conducted a comprehensive ablation study by systematically varying $\tau'$ from 1.0 to 0.1.
Like the experiential weight ablation, we implement SeqXD, S-XKD, and G-XKD under T5-large$\rightarrow$T5-small setting with different $\tau'$ configurations on the XSum dataset.
As demonstrated in Figure \ref{fig:tau-ablation}, all $\mathcal{X}$-KD methods achieve stable performances with respect to variations in the Boltzmann temperature and surpass the corresponding non-experiential baselines (i.e., the dashed lines with corresponding colors.).

\section{Experiment Details}
\label{appendix:hyperparam}

\subsection{Supervised Fine-Tuning}

Like GKD, we perform supervised fine-tuning for all student models and white-box teacher models on all three datasets.

\textbf{Hardware.}
All SFTs are performed on 4 NVIDIA TITAN XP GPUs.
For T5-small and T5-base, we use data parallel and fine-tune all parameters.
For T5-large, we use model parallel and also fine-tune all parameters.
For T5-xl, to save the GPU memory and speed up training, we adopt QLoRA \cite{dettmers2023qlora}, which quantizes the model in 8-bit and only fine-tunes the LoRA \cite{hu2021lora} adapters instead of all parameters.

\textbf{Hyperparameters.}
Table \ref{tab:hyperparam-t5-small-base-sft} shows the SFT hyperparameters for T5-small and T5-base.
For T5-large, we downscale the number of training steps by a factor of 0.4 and downscale the learning rate by factors of 0.05, 0.05, and 0.1 for three tasks respectively, as shown in 
For T5-xl, we further downscale the training steps by factors of 1/8, 1/8, and 1/10 and downscale the learning rates by factors of 1/2, 1/2, and 1/3, as shown in Table \ref{tab:hyperparam-t5-xl-sft}.
The batch sizes are also downscaled from 32 to 4.

\subsection{Distillation to T5-small Student}

\textbf{Hardware.}
Distillation to T5-small student involves large$\rightarrow$small in the primary setup and base$\rightarrow$small and xl$\rightarrow$small in the scalability study of additional experiments.
For base$\rightarrow$small and large$\rightarrow$small, we simply use data parallel and perform training on 4 NVIDIA TITAN XP GPUs.
For xl$\rightarrow$small, for the teacher model (T5-xl), we use quantization and model parallel, and for the student model (T5-small), we use data parallel.
Note that in xl$\rightarrow$small, we put the teacher model on 4 NVIDIA TITAN XP GPUs and put the student model on another 4 NVIDIA TITAN XP GPUs, which costs 8 GPUs in total.
Since every 4 GPUs are on two different servers, we use Ray to build a cluster and ensure efficient communication.

\begin{table}[H]
\caption{Hyperparameters for T5-small and T5-base SFT.}
\label{tab:hyperparam-t5-small-base-sft}
\centering
\setlength\tabcolsep{3pt}
\begin{tabular}{@{}l|lll@{}}
\toprule
\textbf{Hyperparameter}        & \textbf{XSum} & \textbf{WMT} & \textbf{GSM8k} \\ \midrule
Training Steps                 & 20k           & 100k         & 10k            \\
Batch Size                     & 32            & 32           & 32             \\
Eval Split                     & validation    & validation   & test           \\
Learning Rate (LR)             & 0.002         & 0.002        & 0.003          \\
LR Schedule Type               & linear        & linear       & linear         \\
Warmup Steps                   & 2k            & 2k           & 2k             \\
Max Input Length               & 1024          & 64           & 512            \\
Max Output Length              & 64            & 64           & 64             \\ \bottomrule
\end{tabular}
\end{table}

\begin{table}[H]
\caption{Hyperparameters for T5-large SFT.}
\label{tab:hyperparam-t5-large-sft}
\centering
\setlength\tabcolsep{3pt}
\begin{tabular}{@{}l|lll@{}}
\toprule
\textbf{Hyperparameter}        & \textbf{XSum} & \textbf{WMT} & \textbf{GSM8k} \\ \midrule
Training Steps                 & 8k            & 40k          & 10k            \\
Batch Size                     & 32            & 32           & 32             \\
Eval Split                     & validation    & validation   & test           \\
Learning Rate (LR)             & 0.0001        & 0.0001       & 0.0003         \\
LR Schedule Type               & linear        & linear       & linear         \\
Warmup Steps                   & 2k            & 2k           & 2k             \\
Max Input Length               & 1024          & 64           & 512            \\
Max Output Length              & 64            & 64           & 64             \\ \bottomrule
\end{tabular}
\end{table}

\begin{table}[H]
\caption{Hyperparameters for T5-xl SFT.}
\label{tab:hyperparam-t5-xl-sft}
\centering
\setlength\tabcolsep{3pt}
\begin{tabular}{@{}l|lll@{}}
\toprule
\textbf{Hyperparameter}        & \textbf{XSum} & \textbf{WMT} & \textbf{GSM8k} \\ \midrule
Training Steps                 & 1k            & 5k           & 1k             \\
Batch Size                     & 4             & 4            & 4              \\
Eval Split                     & validation    & validation   & test           \\
Learning Rate (LR)             & 0.00005       & 0.00005      & 0.0001         \\
LR Schedule Type               & linear        & linear       & linear         \\
Warmup Steps                   & 0             & 0            & 0              \\
Max Input Length               & 1024          & 64           & 512            \\
Max Output Length              & 64            & 64           & 64             \\ \bottomrule
\end{tabular}
\end{table}

\textbf{Hyperparameters.}
For three different model settings, we use the same hyperparameter configuration, as shown in Table \ref{tab:hyperparam-distill-to-t5-small-base}.
The \textit{Sequence-Level} section contains the decoding hyperparameters for the teacher model.
The general hyperparameters are identical to those in T5-small SFT (Table \ref{tab:hyperparam-t5-small-base-sft}).
Note that the batch size is only applied to gradient descent of the student model.
The no-gradient forward through the teacher model uses a smaller batch size.

\begin{table}[H]
\caption{Hyperparameters for distillations to T5-small and T5-base.}
\label{tab:hyperparam-distill-to-t5-small-base}
\centering
\setlength\tabcolsep{3pt}
\begin{tabular}{@{}l|lll@{}}
\toprule
\textbf{Hyperparameter}        & \textbf{XSum} & \textbf{WMT} & \textbf{GSM8k} \\ \midrule
\textit{General}               & \textbf{}     & \textbf{}    & \textbf{}      \\
Training Steps                 & 20k           & 100k         & 10k            \\
Batch Size                     & 32            & 32           & 32             \\
Eval Split                     & validation    & validation   & test           \\
Learning Rate (LR)             & 0.002         & 0.002        & 0.001          \\
LR Schedule Type               & linear        & linear       & linear         \\
Warmup Steps                   & 2k            & 2k           & 2k             \\
Max Input Length               & 1024          & 64           & 512            \\
Max Output Length              & 64            & 64           & 64             \\ \midrule
\textit{GKD}                   &               &              &                \\
On-Policy Weight $\alpha$      & 0.5           & 0.5          & 0.5            \\
JS Divergence $\beta$          & 0.5           & 0.5          & 0.5            \\ \midrule
\textit{Sequence-Level}        &               &              &                \\
Sampling Temperature           & 1.0           & 1.0          & 1.0            \\
Top-p                          & 0.95          & 0.95         & 0.95           \\ \midrule
\textit{Experiential Learning} &               &              &                \\
Discount Factor $\gamma$       & 1.0           & 1.0          & 1.0            \\
Experiential Weight $\lambda$  & 0.001         & 0.001        & 0.001          \\
Prior Distribution $p(R)$      & Gaussian      & Gaussian     & Gaussian       \\ \bottomrule
\end{tabular}
\end{table}

\subsection{Distillation to T5-base Student}

\textbf{Hardware.}
Distillation to T5-base student involves large$\rightarrow$base in the primary setup and xl$\rightarrow$base in the scalability study of additional experiments.
For large$\rightarrow$base, we simply use data parallel and perform training on 4 NVIDIA TITAN XP GPUs.
For xl$\rightarrow$base, we explicitly separate the teacher and the student, similar to xl$\rightarrow$small.

\textbf{Hyperparameters.}
The hyperparameter configuration for distillation to T5-base is identical to that for distillation to T5-small, as shown in Table \ref{tab:hyperparam-distill-to-t5-small-base}.

\subsection{Distillation to T5-large Student}

\textbf{Hardware.}
Distillation to T5-large student involves black-box distillation from Llama-2-7b to T5-large in the primary setup and xl$\rightarrow$large in the scalability study of additional experiments.
The black-box distillation consists of generating black-box teacher knowledge using Llama-2-7b and fine-tuning T5-large.
We quantize Llama-2-7b with 4-bit quantization and put it on 4 NVIDIA TITAN XP GPUs while putting T5-large on another 4 NVIDIA TITAN XP GPUs with model parallel.
For xl$\rightarrow$large, we also explicitly separate the teacher and the student, similar to xl$\rightarrow$small and xl$\rightarrow$base.

\textbf{Hyperparameters.}
The hyperparameters for distillations to T5-large are shown in Table \ref{tab:hyperparam-distill-to-t5-large}.
The general hyperparameters are identical to those in T5-large SFT (Table \ref{tab:hyperparam-t5-large-sft}).
The decoding hyperparameters in \textit{Sequence-Level} section are applied to both black-box and white-box distillations.

\begin{table}[H]
\caption{Hyperparameters for distillations to T5-large.}
\label{tab:hyperparam-distill-to-t5-large}
\centering
\setlength\tabcolsep{3pt}
\begin{tabular}{@{}l|lll@{}}
\toprule
\textbf{Hyperparameter}        & \textbf{XSum} & \textbf{WMT} & \textbf{GSM8k} \\ \midrule
\textit{General}               & \textbf{}     & \textbf{}    & \textbf{}      \\
Training Steps                 & 8k            & 40k          & 10k            \\
Batch Size                     & 32            & 32           & 32             \\
Eval Split                     & validation    & validation   & test           \\
Learning Rate (LR)             & 0.0001        & 0.0001       & 0.0003         \\
LR Schedule Type               & linear        & linear       & linear         \\
Warmup Steps                   & 2k            & 2k           & 2k             \\
Max Input Length               & 1024          & 64           & 512            \\
Max Output Length              & 64            & 64           & 64             \\ \midrule
\textit{GKD}                   &               &              &                \\
On-Policy Weight $\alpha$      & 0.5           & 0.5          & 0.5            \\
JS Divergence $\beta$          & 0.5           & 0.5          & 0.5            \\ \midrule
\textit{Sequence-Level}        &               &              &                \\
Sampling Temperature           & 1.0           & 1.0          & 1.0            \\
Top-p                          & 0.95          & 0.95         & 0.95           \\ \midrule
\textit{Experiential Learning} &               &              &                \\
Discount Factor $\gamma$       & 1.0           & 1.0          & 1.0            \\
Experiential Weight $\lambda$  & 0.001         & 0.001        & 0.001          \\
Prior Distribution $p(R)$      & Gaussian      & Gaussian     & Gaussian       \\ \bottomrule
\end{tabular}
\end{table}

%%%%%%%%%%%%%%%%%%%%%%%%%%%%%%%%%%%%%%%%%%%%%%%%%%%%%%%%%%%%%%%%%%%%%%%%%%%%%%%
%%%%%%%%%%%%%%%%%%%%%%%%%%%%%%%%%%%%%%%%%%%%%%%%%%%%%%%%%%%%%%%%%%%%%%%%%%%%%%%

\end{document}